\documentclass{article}
\usepackage[T1]{fontenc}
\usepackage[utf8]{inputenc}
\usepackage{colm2024_conference}
\usepackage[english]{babel}
\usepackage{amsmath,amssymb}
\usepackage{graphicx}
\usepackage{booktabs,multirow,array,tabularx}
\usepackage{colortbl}
\usepackage{enumitem}
\usepackage{placeins}
\usepackage{fvextra}
\usepackage{algorithm}
\usepackage{algpseudocode}
\usepackage{caption,subcaption}
\usepackage{xurl}
\usepackage{hyperref}
\definecolor{darkblue}{RGB}{0,45,95}
\hypersetup{colorlinks=true,citecolor=darkblue,linkcolor=darkblue,urlcolor=darkblue,
  pdftitle={Qwen-Image-HD: Autoregressive Diffusion for High Definition Image Generation},
  pdfauthor={},pdfsubject={High-resolution text-to-image generation through local training and blockwise autoregression}}
\usepackage[capitalize,noabbrev]{cleveref}
\makeatletter
\providecommand{\theHALG@line}{}
\renewcommand{\theHALG@line}{\arabic{algorithm}.\arabic{ALG@line}}
\makeatother
\graphicspath{{figures/}}
\setlist[itemize]{leftmargin=*,itemsep=2pt,topsep=3pt}
\newcommand{\ours}{Qwen-Image-HD}
\newcommand{\best}[1]{\textbf{#1}}
\newcommand{\second}[1]{\underline{#1}}

\newcommand{\R}{\mathbb{R}}
\title{Qwen-Image-HD: Autoregressive Diffusion\\for High Definition Image Generation}
\author{%
\mbox{Zihao Liu}, \mbox{Zekai Zhang}, \mbox{Shengming Yin}, \mbox{Zikai Zhou}, \mbox{Donghao Zhou},\\[2pt]
\mbox{Chenyang Zhu}, \mbox{Lihan Jiang}, \mbox{Ningyuan Tang}, \mbox{Bingxuan Zhao}, \mbox{Yanran Zhang},\\[2pt]
\mbox{Yan Shu}, \mbox{Zhening Liu}, \mbox{Jiahao Li}, \mbox{Jie Zhang}, \mbox{Kaiyuan Gao}, \mbox{Kun Yan},\\[2pt]
\mbox{Qihang Ai}, \mbox{Rui Song}, \mbox{Shuo Cao}, \mbox{Xiao Xu}, \mbox{Xiaoyue Chen}, \mbox{Yuxiang Chen},\\[2pt]
\mbox{Zhendong Wang}, \mbox{Zhouyang Li}, \mbox{Bo Zheng}, \mbox{Jun Zhou}, \mbox{Xiu Li}, \mbox{Chenfei Wu\thanks{Corresponding author.}}}

\date{November 24, 2026}

\begin{document}
\maketitle
\begin{abstract}
Fine-tuning a generator on a complete high\babelhyphen{hard}resolution canvas requires long image sequences and substantial computation, yet local regions can still exhibit unnatural repetitive stripes when training resources are limited. We introduce \ours{}, which adapts Qwen-Image 2.0 to high\babelhyphen{hard}resolution synthesis through local training and blockwise autoregressive generation. Each prediction covers a target tile at 1K scale, conditioned on high\babelhyphen{hard}resolution upper and left neighbors and a compressed historical thumbnail. These conditions provide adjacent fine-scale structures and an overview of the scene generated so far. To integrate these complementary cues, we map both forms of history and the target to a shared canvas coordinate system to express their spatial relationships and scale differences. On Qwen-Image-Bench at 4K, \ours{} achieves stronger overall evaluation performance than direct 4K generation with FLUX-3-Image-Klein. Compared with full-image 4K supervised fine-tuning of the same backbone, it achieves better generation quality with lower training compute and higher inference efficiency, while exhibiting fewer unnatural repetitive stripes in original-resolution comparisons. We further introduce a Wavefront\babelhyphen{hard}AutoRegressive inference schedule that generates tiles on the same anti-diagonal concurrently across GPUs to reduce single-image latency.
\end{abstract}

\par\vspace{-3pt}\noindent\hfill
\begin{minipage}{\dimexpr\linewidth-2\leftmargini\relax}
\centering
\includegraphics[width=\linewidth]{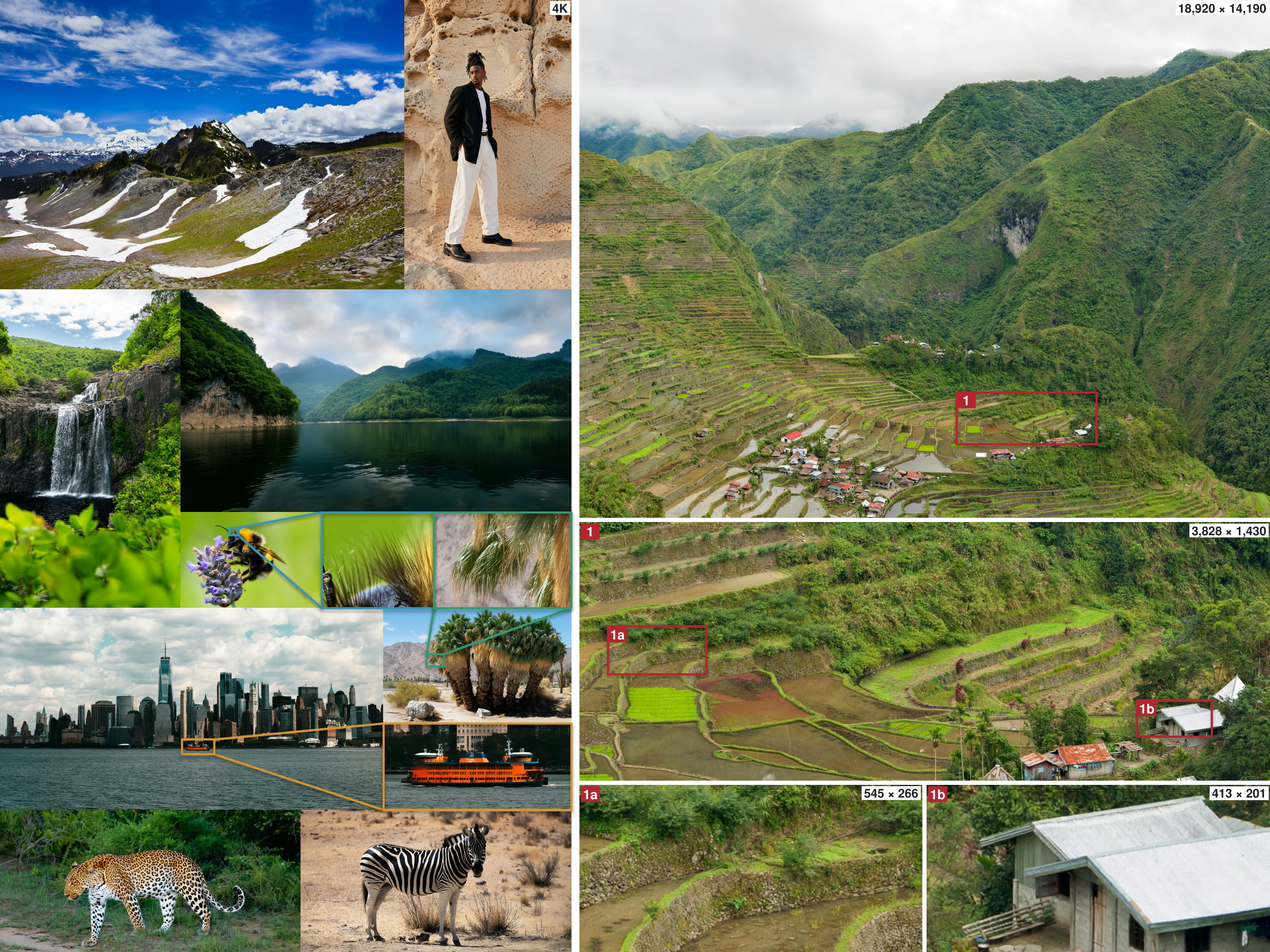}
\captionsetup{font=small,skip=3pt,hypcap=false}
\captionof{figure}{\textbf{Images generated by \ours{}.} Left: multiple 4K examples. Right: a single 16K image with enlarged details.}
\label{fig:teaser}
\end{minipage}\hfill\null\par

\clearpage
\section{Introduction}
\label{sec:introduction}

High\babelhyphen{hard}definition text-to-image generation requires coherent global structure and fine local detail. Recent work~\citep{zhang2025diffusion,ye2026ultraflux} improves 4K synthesis by adapting pretrained generators through post-training with specialized data, representations, and objectives. However, training on complete high\babelhyphen{hard}resolution images requires processing long image sequences at every update. As the canvas grows, this increases the cost of adapting the generator, limiting the number of updates possible within a given budget. Coherent composition alone also does not ensure that individual regions contain convincing detail. This motivates a central question: how can a pretrained generator acquire high\babelhyphen{hard}definition synthesis capabilities efficiently while preserving both scene structure and local detail?

Blockwise prediction shortens the target sequence, but local targets still need context to form a coherent image. NUWA-Infinity~\citep{liang2022nuwa} combines patch- and token-level autoregression, propagating historical information through cached neighborhood features. ACDiT~\citep{hu2024acdit} combines blockwise diffusion with skip-causal attention and KV caching, while MADFormer~\citep{chen2025madformer} separates autoregressive conditioning from iterative diffusion refinement across model layers. Although both reuse historical computations, their retained context grows with the preceding sequence. Reducing target size therefore does not by itself bound the cost of conditioning on a developing canvas. We seek a history representation that supplies nearby detail and an explicit broader overview without retaining every preceding high\babelhyphen{hard}resolution token.

We introduce \ours{}, which adapts Qwen-Image 2.0~\citep{zhao2026qwen} through 1K tile prediction with a bounded, spatially organized history to generate high\babelhyphen{hard}definition images (\cref{fig:teaser}). A fixed-size historical thumbnail summarizes completed canvas regions, while boundary regions of the upper and left neighbors retain fine structures lost during downsampling. Unfilled regions remain white, so the thumbnail represents only available history during both training and generation. Its token count stays fixed as more canvas regions are completed, providing an expanding scene overview without accumulating high\babelhyphen{hard}resolution history. These conditions serve complementary roles: the thumbnail conveys broader scene organization, and the neighbors constrain continuity with adjacent regions. Their different resolutions and spatial extents, however, require the model to distinguish where each condition belongs. We therefore map the thumbnail, neighbors, and target to shared canvas coordinates that preserve their relative positions and scales. This dual-scale representation connects local training targets to the larger scene within a bounded context-token budget.

At inference, Serial\babelhyphen{hard}AutoRegressive sampling uses the same context structure to assemble local predictions into a complete image, updating history after each tile. Sequential sampling, however, increases single-image waiting time even when each prediction is inexpensive. We introduce Wavefront\babelhyphen{hard}AutoRegressive inference to generate tiles on the same anti-diagonal concurrently while preserving upper/left dependencies. Each round uses shared history, which is updated when the round completes. Our implementation combines concurrent tile generation with within-tile parallelism and dynamic GPU allocation to accelerate rounds with fewer active tiles. The schedule applies directly to the Serial\babelhyphen{hard}trained model without additional training, reducing latency through parallel execution with a latency--quality trade-off relative to Serial.

We evaluate adaptation efficiency and generation quality at 4K against training-free resolution extrapolation and training-based high\babelhyphen{hard}resolution methods. A controlled comparison with full-image supervised fine-tuning (SFT) of the same backbone shows that \ours{} reaches higher Q-Judger Overall~\citep{li2026qwen} with lower cumulative training FLOPs under the evaluated budgets and configurations. Original-resolution comparisons and Narrowband Spectral Excess (NSE) on Nature-384 indicate fewer unnatural repetitive stripes than SFT. Controlled comparisons with alternative compact history designs also support the thumbnail representation within our framework. With four GPUs, Wavefront further reduces latency relative to Serial and full-image SFT.

Our main contributions are:
\begin{itemize}
\item \textbf{Dual-scale history and alignment.} We propose a spatially aligned, dual-scale history representation for local adaptation of pretrained image generators, retaining coarse canvas context and fine boundary detail within a bounded context-token budget.
\item \textbf{Wavefront\babelhyphen{hard}AutoRegressive inference.} We develop a Wavefront\babelhyphen{hard}AutoRegressive implementation combining anti-diagonal tile generation, within-tile parallelism, and dynamic GPU allocation to reduce single-image latency.
\item \textbf{Efficient adaptation and generation quality.} We demonstrate strong high\babelhyphen{hard}definition generation quality against existing methods, while controlled same-backbone comparisons show faster convergence and improved quality with lower training compute than full-image SFT.
\end{itemize}

\FloatBarrier
\section{Related Work}
\label{sec:related}

\subsection{Image Generation and Editing}

Recent image foundation models improve generation quality and controllability~\citep{flux2024,wu2025qwen,cao2025hunyuanimage,flux-2-2025,zhao2026qwen,liu2026ernie,flux-3-image-klein-2026}. Visual autoencoders enable higher spatial compression while preserving reconstruction quality~\citep{chen2025deep,yao2025reconstruction,zhang2026qwenvae}, and prompt enhancement makes visual intent more explicit in textual conditions~\citep{wang2025promptenhancer,liu2026ernie,flux-3-image-klein-2026,zhang2026qwenpe}. Image generation synthesizes visual content from input conditions~\citep{zhou2025golden,liu2025controllable,chen2026semhitok}, while image editing enables controlled modification of existing content~\citep{song2026training,liu2026hifi,zhou2024magictailor,zhou2025fireedit}. Within image generation, high-resolution synthesis places additional demands on coherent structure and fine detail across a larger canvas.

\subsection{Training-Free Resolution Extrapolation}

Training-free methods extend pretrained generators to larger canvases while mitigating repetition and structural distortion. For U-Net models, existing approaches adjust convolution dilation~\citep{he2024scalecrafter}, apply frequency filtering~\citep{huang2024fouriscale}, or introduce resolution-aware operations and shifted-window attention~\citep{zhang2024hidiffusion}. Multiscale approaches use low-resolution guidance~\citep{du2024demofusion,kim2025diffusehigh} or coordinate frequency-based sampling~\citep{zhang2024frecas,qiu2025freescale}. Other approaches stabilize patchwise sampling through content conditions and inter-window interaction~\citep{lin2025accdiffusion}, follow low-resolution flow trajectories~\citep{bu2026hiflow}, or combine neighborhood attention with frequency and structural guidance~\citep{koh2026scalediff}.

For diffusion Transformers~\citep{peebles2023scalable}, resolution extrapolation methods adapt positional encoding over denoising~\citep{issachar2026dype}, modulate RoPE by scale and time~\citep{zhaosigma}, or use latent spatial spectra to modulate RoPE contributions~\citep{rajabi2026sega}. Attention-based approaches combine position alignment with hierarchical attention partitioning~\citep{xue2026hrdit}. Our coordinate design instead organizes local conditional inputs, aligning the target, neighbors, and historical thumbnail across regions and scales within a shared canvas frame.

\subsection{High-Resolution Training and Adaptation}

High-resolution training addresses long-sequence cost through compact representations, architectural changes, and efficient adaptation. Sana~\citep{xie2024sana} combines a highly compressed autoencoder with linear attention; UltraPixel~\citep{ren2024ultrapixel} reuses low-resolution semantic representations through implicit neural upsampling and scale-aware normalization, sharing most parameters across resolutions; URAE~\citep{yu2025ultra} studies synthetic training data and selective weight updates.

Data and supervision also shape high-resolution detail. Diffusion-4K~\citep{zhang2025diffusion} introduces Aesthetic-4K and wavelet-based fine-tuning; its extended formulation~\citep{zhang2025ultra} adds SC-VAE and wavelet-based latent fine-tuning. UltraFlux~\citep{ye2026ultraflux} combines multi-aspect-ratio data, Resonance RoPE, a wavelet-based objective, and VAE post-training for native 4K synthesis. Our approach focuses on the prediction target and its historical conditions. Local targets shorten the sequence being denoised, while bounded history retains nearby detail and canvas-level context without accumulating all preceding high-resolution tokens.

\subsection{Blockwise Autoregressive Generation}

Blockwise generation and historical information reuse have established precedents. NUWA-Infinity~\citep{liang2022nuwa} combines patch- and token-level autoregression, propagating history through cached neighborhood features. ACDiT~\citep{hu2024acdit} uses skip-causal attention and KV caching for blockwise diffusion, while MADFormer~\citep{chen2025madformer} separates autoregressive conditioning from diffusion refinement across model layers. Although ACDiT and MADFormer reuse historical computations, their retained context grows with preceding blocks. \ours{} uses a bounded, dual-scale history: a full-canvas thumbnail summarizes completed regions, upper and left neighbor boundaries retain fine detail, and shared canvas coordinates preserve their positions and scales relative to the target. To evaluate alternatives to the thumbnail, our same-backbone controlled comparisons examine Recurrent Memory inspired by NUWA-Infinity's neighborhood reuse and Learned Compression inspired by MADFormer's conditioning--refinement separation. Both use the same history-token budget, neighbor inputs, and canvas alignment. The thumbnail achieves higher Q-Judger Overall than both alternatives and the w/o Historical Thumbnail variant (\cref{sec:ablations}).

\section{Method}
\label{sec:method}

\begin{figure}[!tbp]
\centering
\includegraphics[width=\linewidth]{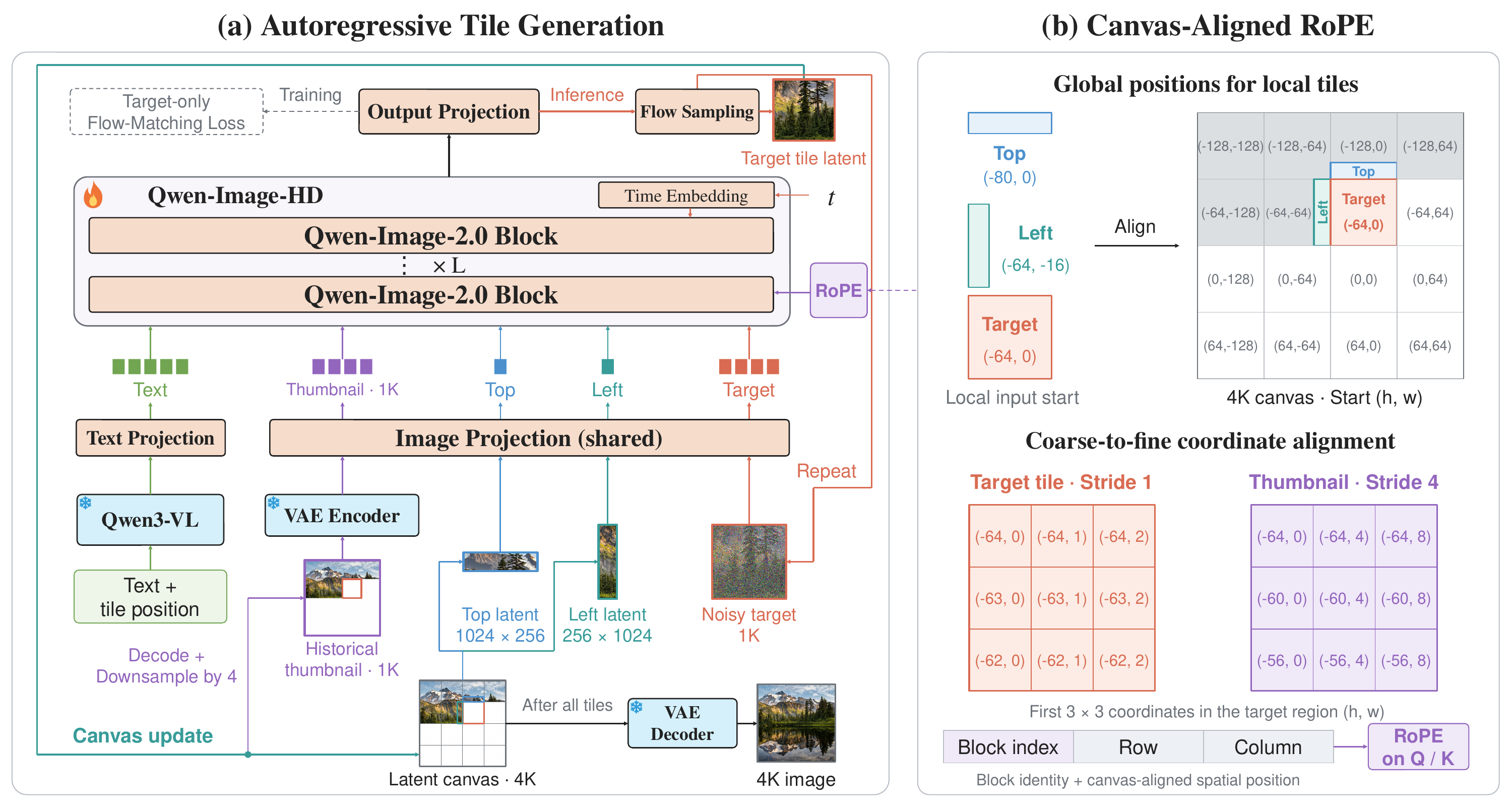}
\caption{\textbf{Architecture and spatial alignment of \ours{}.} Left: the historical thumbnail, upper/left neighbors, and noisy target tile serve as inputs, with separate paths for the training loss and inference updates. The sampled latent is written to the latent canvas and separately decoded and downsampled into the historical thumbnail. Right: spatial relationships and scale-dependent coordinate strides in a shared canvas frame. Image sizes are given in pixels (width $\times$ height).}
\label{fig:framework}
\end{figure}

\ours{} adapts a pretrained image generator through local prediction conditioned on a spatially aligned, dual-scale history. A historical thumbnail retains coarse canvas context, while neighboring boundary latents provide fine detail; shared canvas coordinates align both with the target tile. \Cref{fig:framework} summarizes the history construction, spatial alignment, and canvas updates.

\subsection{Problem Formulation}
\label{sec:formulation}

Given a text condition $c$ and output resolution $H\times W$, we model the conditional generation distribution of an image $x\in\R^{H\times W\times3}$. We specify $R$ rows and $C$ columns and uniformly partition the canvas into $R\times C$ non-overlapping rectangular tiles. When $H$ and $W$ are divisible by $R$ and $C$, respectively, tile height, tile width, and the total number of tiles are
\begin{equation}
B_h=\frac{H}{R},\qquad B_w=\frac{W}{C},\qquad K=RC.
\end{equation}
The grid is chosen according to the output resolution and aspect ratio so that tile dimensions satisfy the encoder constraints and remain at approximately 1K spatial scale. Our square 4K evaluation uses $H=W=4096$ and $R=C=4$, giving $B_h=B_w=1024$. This parameterization also supports rectangular tiles and other grid sizes.

Let $Z=(z_1,\ldots,z_K)$ denote latent tiles in Serial order: left to right within each row, from top to bottom. The conditional distribution factorizes as
\begin{equation}
p_\theta(Z\mid c,H,W)
=\prod_{i=1}^{K}p_\theta\!\left(z_i\mid c_i,\mathcal{N}_i,M_{<i},P_i\right).
\label{eq:factorization}
\end{equation}
Here, $\mathcal{N}_i$ contains the completed upper and left neighbor latents, $M_{<i}$ is the historical thumbnail condition before tile $i$, and $P_i$ contains spatial coordinates and block identifiers.

The tile-level text $c_i=T(c,i,R,C)$ uses a fixed template to prepend the grid size and target row/column to the input text. Training and inference use the same template. In the following example, $r$ and $c$ denote the target row and column indices, with $1\leq r\leq R$ and $1\leq c\leq C$:
\begin{Verbatim}[fontsize=\small,breaklines=true,frame=single,framesep=5pt]
This is one tile of a large image arranged in an R×C grid (row r, column c).
The original caption is: <caption>
\end{Verbatim}

\textbf{Rectified Flow within each tile.}
We model each tile-level conditional distribution using Rectified Flow~\citep{liu2022flow}. A straight interpolation between the target latent and Gaussian noise defines a probability path, and conditional flow matching~\citep{lipman2022flow} learns its velocity field. For a clean target latent $z_i$, standard Gaussian noise $\epsilon\sim\mathcal N(0,I)$, and time $t\in[0,1]$, the intermediate state and target velocity are
\begin{equation}
z_i^t=(1-t)z_i+t\epsilon,\qquad
v_i^t=\frac{\mathrm d z_i^t}{\mathrm d t}=\epsilon-z_i.
\end{equation}
The endpoints $t=0$ and $t=1$ correspond to the clean latent and Gaussian noise. Conditioned on tile-level text, neighbors, the historical thumbnail, and position encodings, the model predicts the target velocity by minimizing
\begin{equation}
\mathcal L(\theta)=\mathbb E_{(x,c,i,\epsilon,t)\sim\mathcal D}\!\left[
\left\|v_\theta\!\left(z_i^t,t;c_i,\mathcal N_i,M_{<i},P_i\right)-v_i^t\right\|_2^2
\right],
\label{eq:loss}
\end{equation}
where $\mathcal D$ denotes the training dataset. Only the target tile is noised and supervised; the neighbors and thumbnail remain clean.

\subsection{Two Scales of Historical Context}
\label{sec:history}

\textbf{High-resolution neighbors.}
The model conditions on completed upper and left neighbor latents according to the target's position. For a $1024\times1024$ target tile, the upper neighbor covers the bottom $1024\times256$ region of the completed tile above, and the left neighbor covers the rightmost $256\times1024$ region of the completed tile to the left, with dimensions given as width $\times$ height in pixels. For general tile sizes, the crop depths are $d_h=\min(256,B_h)$ pixels for the upper neighbor and $d_w=\min(256,B_w)$ pixels for the left neighbor, using the full available extent when it is below 256 pixels. The corresponding latents are extracted from the stored tile latents. The top-left tile has no neighbors. For the remaining tiles, first-row tiles use only the left neighbor, first-column tiles use only the upper neighbor, and interior tiles use both. These neighbors directly adjoin the target boundaries, providing fine structures, texture directions, and local geometry for continuous generation across tiles.

\textbf{Historical thumbnail.}
Completed pixel-space regions are downsampled and written into a thumbnail covering the full canvas; current and future regions remain white. In particular, the thumbnail is entirely white for the first tile. A frozen VAE encodes the thumbnail as a latent condition. For a $4096\times4096$ canvas, the thumbnail is $1024\times1024$; each new tile is downsampled by 4 along both axes before insertion. This compact representation retains subject distribution, layout, and color information from the generated scene.
Training history is constructed from ground-truth regions preceding the target tile, whereas inference history uses completed model outputs.

\textbf{Context budget.}
Let the target contain $n$ image tokens, let at most $k$ neighbors with $p$ tokens each be used, and let the thumbnail contain $m$ tokens. The image input length per model call satisfies
\begin{equation}
L_{\mathrm{image}}\leq n+kp+m.
\label{eq:budget}
\end{equation}
For square 4K generation, $n=m=4096$, $p=1024$, and $k=2$, giving an upper bound of 10,240 image tokens. Direct full-image 4K generation processes 65,536 image tokens, 6.4 times this bound. Text tokens are excluded from both counts. With fixed target and neighbor dimensions, neighbor count, and thumbnail budget, the bound is independent of canvas size. Generating the complete image still requires storing the canvas and sampling every tile, so total computation and canvas storage grow with output size.

\subsection{Canvas Coordinates and Cross-Scale Alignment}
\label{sec:coordinates}

The pretrained Qwen-Image 2.0~\citep{zhao2026qwen} backbone uses rotary position embeddings (RoPE)~\citep{su2024roformer} with three axes encoding image-block identity, row, and column. Separate image inputs use independently centered spatial coordinates. Applied directly to tiles, this convention assigns identical coordinates to local positions in different tiles, while physically adjacent tokens across boundaries may have large coordinate differences. This obscures spatial adjacency and true relative displacement. We therefore retain the block-identifier axis but replace row and column coordinates with positions on the full canvas. Tokens within each input block share an identifier, and different blocks receive increasing identifiers in sequence order, distinguishing the historical thumbnail, neighbors, and target.

\textbf{Neighbor and target tiles.}
Let $f$ be the pixel-to-latent downsampling factor. The latent canvas is $G_h\times G_w=(H/f)\times(W/f)$, and each complete grid tile is $b_h\times b_w=(B_h/f)\times(B_w/f)$. For a one-indexed grid tile $(r,c)$ and zero-indexed position $(u,v)$ within that complete tile, the canvas coordinates are
\begin{equation}
P_{\mathrm{tile}}^{r,c}(u,v)
=\left((r-1)b_h+u+o_h,\;(c-1)b_w+v+o_w\right),
\label{eq:tile-coordinates}
\end{equation}
where $o_h$ and $o_w$ are the offset parameters that center the full latent canvas along the height and width axes, respectively. For a neighbor, $(r,c)$ identifies its source grid tile, and $(u,v)$ retains the position within that complete tile after cropping. The upper neighbor includes the row offset of the bottom crop, while the left neighbor includes the column offset of the rightmost crop. Neighbors and the target therefore use the same mapping to their actual canvas positions, preserving their spatial adjacency.

\textbf{Historical thumbnail.}
The historical thumbnail shares the canvas aspect ratio and its $R\times C$ partition. Let its latent dimensions be $g_h\times g_w$ and the scale ratio be $s=G_h/g_h=G_w/g_w$. Its coordinates are
\begin{equation}
P_{\mathrm{thumb}}(u,v)=\left(su+o_h,\;sv+o_w\right).
\label{eq:thumbnail-coordinates}
\end{equation}
In our square 4K setting, $f=16$, the latent canvas is $256\times256$, and both a tile and the thumbnail are $64\times64$. Tiles use unit coordinate strides, and the thumbnail uses stride $s=4$. For example, the target at row 2, column 3 has top-left coordinates $(-64,0)$. Consecutive tile row coordinates are $-64,-63,-62$, while the corresponding thumbnail coordinates are $-64,-60,-56$. Coarse tokens thus cover larger spatial intervals while sharing canvas positions with local tiles.

\par\noindent
\begin{minipage}{\linewidth}
\setlength{\intextsep}{3pt}
\begin{algorithm}[H]
\caption{Local conditional training}
\label{alg:training}
\small
\begin{algorithmic}[1]
\Require Training dataset $\mathcal D$; grid rows $R$, columns $C$; $S$ updates
\State Initialize $\theta$ from the pretrained generator; freeze text encoder and VAE
\For{$s=1,\ldots,S$}
  \State Sample a batch $\mathcal B$ of tuples $(x,c,i)$ from $\mathcal D$
  \ForAll{$(x,c,i)\in\mathcal B$}
    \State $H,W\gets$ image dimensions; $B_h\gets H/R$; $B_w\gets W/C$
    \State Partition $x$ into an $R\times C$ grid of $B_h\times B_w$ tiles in Serial order
    \State $Z^{\mathrm{gt}}\gets\operatorname{VAEEncode}(x)$; $z_i\gets$ target latent tile $i$
    \If{the target is the top-left tile}
      \State $\mathcal N_i\gets\varnothing$
    \ElsIf{the target is in the first row}
      \State $\mathcal N_i\gets$ rightmost $d_w/f$ columns of the left tile in $Z^{\mathrm{gt}}$
    \ElsIf{the target is in the first column}
      \State $\mathcal N_i\gets$ bottom $d_h/f$ rows of the upper tile in $Z^{\mathrm{gt}}$
    \Else
      \State $\mathcal N_i\gets$ corresponding upper and left neighbor crops from $Z^{\mathrm{gt}}$
    \EndIf
    \State Initialize a white thumbnail $m$; insert downsampled image tiles $j<i$
    \State $M\gets\operatorname{VAEEncode}(m)$; $c_i\gets T(c,i,R,C)$; $e_i\gets\operatorname{TextEncode}(c_i)$
    \State $P_i\gets$ canvas coordinates and block identifiers of all image inputs
    \State Sample denoising time $t\in(0,1)$ and noise $\epsilon\sim\mathcal N(0,I)$
    \State $z_i^t\gets(1-t)z_i+t\epsilon$
    \State $\widehat v\gets v_\theta(z_i^t,t;e_i,\mathcal N_i,M,P_i)$
    \State $\ell_i\gets\operatorname{MSE}(\widehat v,\epsilon-z_i)$
  \EndFor
  \State Backpropagate the mean loss over $\mathcal B$ and update $\theta$
\EndFor
\State \Return Trained generator parameters $\theta$
\end{algorithmic}
\end{algorithm}

\begin{algorithm}[H]
\caption{Serial autoregressive generation}
\label{alg:inference}
\small
\begin{algorithmic}[1]
\Require Text $c$; canvas $H\times W$; grid rows $R$, columns $C$
\State $B_h\gets H/R$; $B_w\gets W/C$
\State Initialize an empty latent canvas $Z$ and white historical thumbnail $m$
\ForAll{tiles $i$ in Serial order}
  \If{the current tile is the top-left tile}
    \State $\mathcal N_i\gets\varnothing$
  \ElsIf{the current tile is in the first row}
    \State $\mathcal N_i\gets$ rightmost $d_w/f$ columns of the completed left tile in $Z$
  \ElsIf{the current tile is in the first column}
    \State $\mathcal N_i\gets$ bottom $d_h/f$ rows of the completed upper tile in $Z$
  \Else
    \State $\mathcal N_i\gets$ corresponding upper and left neighbor crops from completed tiles in $Z$
  \EndIf
  \State $M\gets\operatorname{VAEEncode}(m)$; $c_i\gets T(c,i,R,C)$
  \State $P_i\gets$ canvas coordinates of the neighbors, target, and thumbnail
  \State Sample $z_i$ from noise using conditional Rectified Flow with $(c_i,\mathcal N_i,M,P_i)$
  \State Write $z_i$ into the corresponding region of $Z$
  \State Downsample $\operatorname{VAEDecode}(z_i)$ and insert it into $m$
\EndFor
\State \Return VAE decoding of the complete latent canvas $Z$
\end{algorithmic}
\end{algorithm}
\end{minipage}\par

\subsection{Training and Autoregressive Inference}
\label{sec:algorithms}

\textbf{Training.}
Each training sample consists of an image, text, and a target tile index. According to the target's position in the Serial order, upper and left neighbors are extracted from the ground-truth latents using the crop depths defined in \cref{sec:history}, corresponding to $d_h/f$ latent rows and $d_w/f$ latent columns, respectively. Preceding image regions are written to the thumbnail, completing the causal conditions. The model predicts the noised target and updates the generation Transformer using conditional flow matching. The mean squared error in \cref{alg:training} averages only over target latent elements.

\textbf{Autoregressive inference.}
Generation begins with a white historical thumbnail and an empty latent canvas. Each step extracts upper and left neighbors from completed tiles in the generated latent canvas using the same boundary crops as in training, encodes the historical thumbnail, constructs text and spatial coordinates, and samples the target from Gaussian noise. The output follows two separate update paths: the latent is written to the complete canvas for neighbor reuse and final decoding, and its decoded pixels are downsampled into the thumbnail (\cref{alg:inference}). After all tiles are complete, VAE decoding converts the assembled latent canvas to the final image.

\subsection{Wavefront Parallel Inference}
\label{sec:wavefront}

\begin{figure}[!t]
\centering
\includegraphics[width=\linewidth]{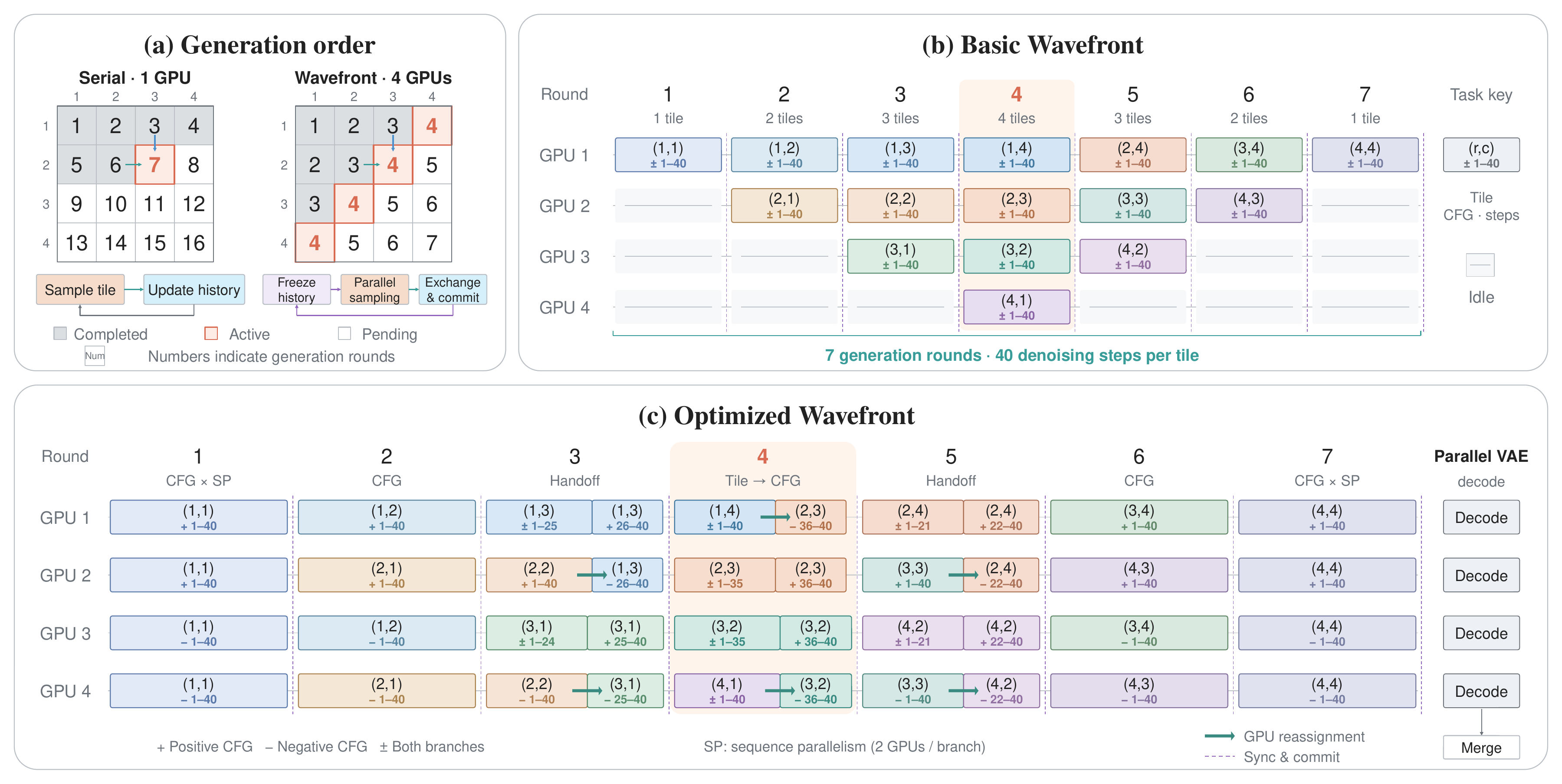}
\caption{\textbf{Generation order and cooperative Wavefront scheduling.} (a) Serial and Wavefront traversal with their history updates. (b) Basic scheduling with one GPU per tile. (c) Optimized scheduling combines tile parallelism, CFG-branch parallelism, within-branch sequence parallelism (SP), GPU task handoffs, and final parallel VAE decoding.}
\label{fig:wavefront}
\end{figure}

\textbf{Basic Wavefront.}
To shorten sequential dependencies, we group tiles on the same anti-diagonal:
\begin{equation}
\mathcal W_d=\{(r,c):r+c-2=d\},\qquad d=0,\ldots,R+C-2.
\end{equation}
Both the upper and left neighbor lie in the preceding round. In the basic Wavefront schedule, all tiles in a round read a fixed history and sample on separate GPUs; their outputs are then exchanged to update the shared canvas. Each GPU holds a complete model replica. An $R\times C$ grid has $R+C-1$ rounds and maximum round width $\min(R,C)$. If fewer GPUs are available than active tiles, each GPU processes its assigned tiles sequentially while the round's history remains fixed.

If a tile takes approximately $\tau$ to sample, round $d$ contains $w_d$ tiles, and $P$ GPUs are available, idealized sampling times using only tile parallelism are
\begin{equation}
T_{\mathrm{serial}}\approx RC\tau,\qquad
T_{\mathrm{wave}}\approx\sum_{d=0}^{R+C-2}\left\lceil\frac{w_d}{P}\right\rceil\tau.
\end{equation}
On four GPUs, sampling multiple tiles of a $4\times4$ canvas can overlap, reducing single-image latency. This estimate describes the basic schedule in \cref{fig:wavefront}(b) and excludes variation in condition length, communication, history updates, and final decoding.

Wavefront history contains completed anti-diagonals, while Serial history contains a prefix of the Serial order, so a location can access different distant history under the two schedules. \Cref{sec:inference-results} fixes the model weights to compare generation quality, single-image latency, and GPU resource use.

\textbf{Optimized Wavefront.}
The basic schedule leaves resources underused when a round has few tiles or their sampling times differ. We therefore combine classifier-free guidance (CFG) branch parallelism with within-branch sequence parallelism, as shown in \cref{fig:wavefront}(c). In a one-tile round, four GPUs split the two CFG branches, with two GPUs per branch. In a two-tile round, two GPUs handle the positive and negative branches of each tile. For rounds containing three or four tiles, a GPU that finishes early takes over one CFG branch of an unfinished tile at a predetermined denoising step, so the remaining steps run cooperatively on two GPUs. Tiles still exchange outputs and update history together at the end of the round. Parallel VAE decoding produces the final image once the full latent canvas is complete.

\FloatBarrier
\section{Experiments}
\label{sec:experiments}

\subsection{Experimental Setup}

We evaluate \ours{} from three perspectives: 4K generation quality, training efficiency, and parallel inference. Comparisons with external methods assess generation quality, while a controlled same-backbone comparison with full-image SFT examines training cost, visual quality, and unnatural repetitive stripes. We further study historical conditions and spatial alignment through ablations, and evaluate the latency, generation quality, and GPU-normalized throughput of Wavefront inference.

\textbf{Training.}
We initialize the generation Transformer from Qwen-Image 2.0~\citep{zhao2026qwen} and train all its parameters, while freezing the text encoder and VAE. We use AdamW~\citep{loshchilov2017decoupled} with $(\beta_1,\beta_2)=(0.9,0.95)$, a global batch size of 64, a learning rate of $10^{-5}$, 200 warmup steps, and gradient clipping at 1.0. Our 20k-step training costs 966 GPU$\cdot$h.

\begin{figure}[H]
\centering
\includegraphics[width=0.75\linewidth]{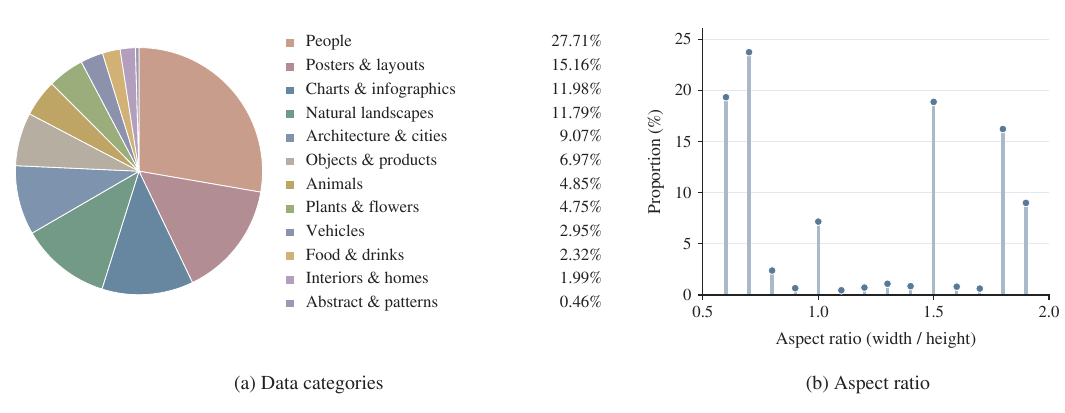}
\caption{\textbf{Illustration of training-data categories and aspect ratios.}}
\label{fig:training-data}
\end{figure}

\textbf{Training data.}
We collect 4K images spanning portraits, natural landscapes, animals, architecture, and urban scenes, with varied resolutions and aspect ratios. \Cref{fig:training-data} illustrates the content categories and aspect ratios.

\vspace{-6pt}

\textbf{Benchmarks and generation settings.}
We evaluate image generation at $4096\times4096$ resolution on Aesthetic-4K~\citep{zhang2025diffusion}, DrawBench~\citep{saharia2022photorealistic}, and Qwen-Image-Bench~\citep{li2026qwen}, with 195, 200, and 1,000 prompts, respectively. \ours{} uses a $4\times4$ grid, 40 sampling steps per tile, and classifier-free guidance (CFG)~\citep{ho2022classifier} scale 4; external methods retain their published sampling configurations. Nature-384, used to diagnose unnatural repetitive stripes, is introduced in \cref{sec:training-results}.

\textbf{Compared methods.}
Aesthetic-4K and DrawBench include 21 external methods covering training-free resolution extrapolation and trained high-resolution generation or adaptation. Qwen-Image-Bench focuses on trained generators, including FLUX-3-Image-Klein-Base and FLUX-3-Image-Klein. External methods retain their own backbones. All three benchmarks also include full-image SFT, which shares our pretrained backbone, training data sources, and training hyperparameters. SFT's 6k-step training costs 4,426 GPU$\cdot$h.

\textbf{Evaluation metrics.}
We use the official Q-Judger~\citep{li2026qwen} on all three benchmarks, reporting Quality, Aesthetics, Alignment, Real-world Fidelity, Creative Generation, and Overall. On Aesthetic-4K and DrawBench, HPSv3~\citep{ma2025hpsv3} and PickScore~\citep{kirstain2023pick} assess image--text preference, while ArtiMuse~\citep{cao2026artimuse} assesses visual quality and aesthetics.

\subsection{Main Results}
\label{sec:main-results}

\textbf{Qwen-Image-Bench.}
In \cref{tab:qib}, \ours{} obtains the highest Quality, Aesthetics, Alignment, Creative Generation, and Overall scores. Its Overall score of 59.44 exceeds FLUX-3-Image-Klein-Base by 1.75 points. Against same-backbone full-image SFT, it scores higher on all five dimensions and improves Overall by 1.17 points. \Cref{sec:ablations} examines the contributions of historical conditions and spatial alignment.

\begin{table}[!htbp]
\centering
\caption{\textbf{4K results on Qwen-Image-Bench.} All scores are higher-is-better. Bold and underlining denote the best and second-best entries in each column.}
\label{tab:qib}
\begingroup
\fontsize{8.5}{10.2}\selectfont
\setlength{\tabcolsep}{0pt}
\renewcommand{\arraystretch}{1.10}
\setlength{\aboverulesep}{3pt}
\setlength{\belowrulesep}{3pt}
\begin{tabular}{@{}>{\raggedright\arraybackslash}m{.340\linewidth}>{\centering\arraybackslash}m{.095\linewidth}>{\centering\arraybackslash}m{.117\linewidth}>{\centering\arraybackslash}m{.121\linewidth}>{\centering\arraybackslash}m{.118\linewidth}>{\centering\arraybackslash}m{.121\linewidth}>{\centering\arraybackslash}m{.088\linewidth}@{}}
\toprule[0.8pt]
\textbf{Method} & \textbf{Quality}\,$\uparrow$ & \textbf{Aesthetics}\,$\uparrow$ & \textbf{Alignment}\,$\uparrow$ & \shortstack[c]{\textbf{Real-World}\\\textbf{Fidelity}}\,$\uparrow$ & \shortstack[c]{\textbf{Creative}\\\textbf{Generation}}\,$\uparrow$ & \textbf{Overall}\,$\uparrow$ \\
\midrule[0.4pt]
\rowcolor{black!6}
\multicolumn{7}{@{}l@{}}{\hspace{4pt}\rule[-3pt]{0pt}{13pt}\textbf{External Methods}} \\
\addlinespace[2pt]
UltraPixel~\citep{ren2024ultrapixel} & 51.43 & 55.29 & 41.71 & 40.35 & 39.54 & 47.23 \\
Sana~\citep{xie2024sana} & 46.66 & 52.36 & 46.29 & 40.34 & 38.55 & 46.17 \\
URAE~\citep{yu2025ultra} & 46.59 & 51.61 & 46.42 & 40.54 & 41.44 & 46.41 \\
Diffusion-4K~\citep{zhang2025diffusion} & 44.39 & 45.43 & 38.97 & 39.71 & 33.01 & 41.13 \\
UltraFlux~\citep{ye2026ultraflux} & 51.09 & 56.24 & 49.55 & 42.30 & 43.78 & 50.00 \\
FLUX-3-Image-Klein-Base~\citep{flux-3-image-klein-2026} & 53.93 & 62.57 & \second{60.35} & \best{48.59} & 58.19 & 57.69 \\
FLUX-3-Image-Klein~\citep{flux-3-image-klein-2026} & 54.48 & 61.81 & 59.95 & \second{47.80} & 57.67 & 57.37 \\
\addlinespace[4pt]
\rowcolor{black!6}
\multicolumn{7}{@{}l@{}}{\hspace{4pt}\rule[-3pt]{0pt}{13pt}\textbf{Same-Backbone Comparison}} \\
\addlinespace[2pt]
Qwen-Image 2.0 SFT & \second{55.25} & \second{64.54} & 59.35 & 46.49 & \second{60.20} & \second{58.27} \\
\textbf{Qwen-Image-HD} & \best{56.99} & \best{65.79} & \best{60.42} & 47.59 & \best{60.59} & \best{59.44} \\
\bottomrule[0.8pt]
\end{tabular}
\endgroup
\end{table}

\textbf{Aesthetic-4K and DrawBench.}
In \cref{tab:aes4k-drawbench}, \ours{} leads the external methods on HPSv3, PickScore, ArtiMuse, and Q-Judger Overall on both benchmarks. Its Q-Judger Overall exceeds the strongest external result by 2.59 and 1.66 points, respectively. The same-backbone comparison also favors \ours{} on Q-Judger Quality, Aesthetics, Creative Generation, and Overall across both benchmarks.

\begin{table}[!tbp]
\centering
\caption{\textbf{Results on Aesthetic-4K (top) and DrawBench (bottom).} Best and second-best results are bold and underlined, respectively.}
\label{tab:aes4k-drawbench}
\begingroup
\fontsize{8}{9.6}\selectfont
\setlength{\tabcolsep}{0pt}
\renewcommand{\arraystretch}{1.05}
\setlength{\aboverulesep}{2pt}
\setlength{\belowrulesep}{2pt}
\begin{tabular}{@{}>{\raggedright\arraybackslash}m{.310\linewidth}>{\centering\arraybackslash}m{.075\linewidth}>{\centering\arraybackslash}m{.094\linewidth}>{\centering\arraybackslash}m{.091\linewidth}>{\centering\arraybackslash}m{.066\linewidth}>{\centering\arraybackslash}m{.073\linewidth}>{\centering\arraybackslash}m{.077\linewidth}>{\centering\arraybackslash}m{.075\linewidth}>{\centering\arraybackslash}m{.077\linewidth}>{\centering\arraybackslash}m{.062\linewidth}@{}}
\toprule[0.8pt]
\multirow{3}{*}[-3pt]{\textbf{Method}} & \multirow{3}{*}[-3pt]{\textbf{HPSv3}\,$\uparrow$} & \multirow{3}{*}[-3pt]{\textbf{PickScore}\,$\uparrow$} & \multirow{3}{*}[-3pt]{\textbf{ArtiMuse}\,$\uparrow$} & \multicolumn{6}{c}{\textbf{Q-Judger}} \\
\cmidrule(lr){5-10}
 & & & & \multirow{2}{*}{{\fontsize{6.8}{9}\selectfont\textit{Quality}\,$\uparrow$}} & \multirow{2}{*}{{\fontsize{6.8}{9}\selectfont\textit{Aesthetics}\,$\uparrow$}} & \multirow{2}{*}{{\fontsize{6.8}{9}\selectfont\textit{Alignment}\,$\uparrow$}} & {\fontsize{6.8}{9}\selectfont\textit{Real-World}} & {\fontsize{6.8}{9}\selectfont\textit{Creative}} & \multirow{2}{*}{{\fontsize{6.8}{9}\selectfont\textit{Overall}\,$\uparrow$}} \\
 & & & & & & & {\fontsize{6.8}{9}\selectfont\textit{Fidelity}\,$\uparrow$} & {\fontsize{6.8}{9}\selectfont\textit{Generation}\,$\uparrow$} & \\
\midrule[0.4pt]
\rowcolor{darkblue!8}
\multicolumn{10}{c}{\rule[-3pt]{0pt}{13pt}\textbf{Aesthetic-4K}} \\
\addlinespace[2pt]

\rowcolor{black!6}
\multicolumn{10}{@{}l@{}}{\hspace{4pt}\rule[-2pt]{0pt}{11pt}\textbf{Training-Free}} \\
ScaleCrafter~\citep{he2024scalecrafter} & 6.52 & 21.59 & 67.66 & 42.02 & 51.66 & 43.17 & 42.82 & 34.67 & 42.87 \\
DemoFusion~\citep{du2024demofusion} & 8.08 & 22.75 & 61.96 & 36.19 & 44.04 & 48.14 & 42.98 & 39.90 & 42.25 \\
HiDiffusion~\citep{zhang2024hidiffusion} & 7.45 & 22.07 & 65.05 & 39.99 & 55.24 & 45.89 & 43.77 & 33.89 & 43.80 \\
FouriScale~\citep{huang2024fouriscale} & 7.32 & 21.81 & 65.89 & 44.23 & 56.49 & 45.55 & 43.37 & 35.80 & 45.11 \\
DiffuseHigh~\citep{kim2025diffusehigh} & 8.89 & 22.89 & 63.89 & 41.44 & 59.28 & 52.93 & 43.87 & 41.70 & 47.85 \\
FreCaS~\citep{zhang2024frecas} & 9.58 & 22.73 & 68.14 & 48.10 & 61.96 & 51.81 & 44.49 & 41.39 & 49.59 \\
AccDiffusion~\citep{lin2025accdiffusion} & 8.35 & 22.75 & 63.97 & 42.59 & 58.93 & 53.30 & 44.24 & 38.38 & 47.52 \\
FreeScale~\citep{qiu2025freescale} & 10.04 & 22.78 & 63.17 & 43.78 & 57.33 & 52.68 & 42.70 & 39.58 & 47.23 \\
HiFlow~\citep{bu2026hiflow} & 12.47 & 23.38 & 67.70 & 46.37 & 61.96 & 58.55 & 42.98 & 40.82 & 50.14 \\
DyPE~\citep{issachar2026dype} & 1.79 & 20.43 & 58.45 & 24.69 & 32.21 & 34.97 & 41.99 & 18.81 & 30.59 \\
ScaleDiff~\citep{koh2026scalediff} & 12.02 & 23.25 & 66.13 & 44.89 & 54.06 & 53.25 & 43.68 & 40.62 & 47.31 \\
SEGA~\citep{rajabi2026sega} & 6.94 & 21.71 & 63.85 & 33.52 & 47.39 & 46.49 & 44.32 & 30.26 & 40.44 \\
SigMa~\citep{zhaosigma} & 8.78 & 22.18 & 70.58 & 45.59 & 55.95 & 52.34 & 46.95 & 37.82 & 47.77 \\
HRDiT~\citep{xue2026hrdit} & 12.73 & 23.48 & 68.46 & 44.84 & 60.50 & 57.87 & 43.86 & 42.79 & 49.97 \\
\addlinespace[2pt]
\rowcolor{black!6}
\multicolumn{10}{@{}l@{}}{\hspace{4pt}\rule[-2pt]{0pt}{11pt}\textbf{Training-Based}} \\
UltraPixel~\citep{ren2024ultrapixel} & 10.86 & 23.54 & 70.28 & 55.29 & 64.97 & 55.37 & 46.70 & 45.08 & 53.52 \\
Sana~\citep{xie2024sana} & 10.80 & 23.15 & 64.71 & 45.17 & 61.95 & 58.03 & 42.72 & 42.33 & 50.05 \\
URAE~\citep{yu2025ultra} & 10.06 & 22.36 & 66.95 & 44.74 & 57.97 & 56.37 & 43.72 & 38.63 & 48.31 \\
Diffusion-4K~\citep{zhang2025diffusion} & 8.29 & 21.67 & 62.03 & 44.64 & 54.24 & 51.78 & 43.83 & 35.55 & 46.04 \\
UltraFlux~\citep{ye2026ultraflux} & 12.48 & 23.54 & \second{70.88} & 52.55 & 65.92 & 61.27 & 44.66 & 45.21 & 53.97 \\
FLUX-3-Image-Klein~\citep{flux-3-image-klein-2026} & 12.05 & 23.30 & 70.21 & 54.80 & 65.82 & 62.82 & \second{47.36} & 46.79 & 55.52 \\
FLUX-3-Image-Klein-Base~\citep{flux-3-image-klein-2026} & 11.38 & 23.04 & 69.84 & 54.78 & 66.60 & 63.46 & \best{47.81} & 48.46 & 56.24 \\
\addlinespace[2pt]
\rowcolor{black!6}
\multicolumn{10}{@{}l@{}}{\hspace{4pt}\rule[-2pt]{0pt}{11pt}\textbf{Same-Backbone Comparison}} \\
Qwen-Image 2.0 SFT & \second{12.89} & \second{23.75} & 70.73 & \second{56.17} & \second{71.58} & \best{64.21} & 47.21 & \second{48.74} & \second{57.60} \\
\textbf{Qwen-Image-HD} & \best{13.30} & \best{23.85} & \best{72.79} & \best{58.81} & \best{72.28} & \second{64.17} & 46.50 & \best{52.37} & \best{58.83} \\
\midrule[0.8pt]
\addlinespace[4pt]
\rowcolor{darkblue!8}
\multicolumn{10}{c}{\rule[-3pt]{0pt}{13pt}\textbf{DrawBench}} \\
\addlinespace[2pt]

\rowcolor{black!6}
\multicolumn{10}{@{}l@{}}{\hspace{4pt}\rule[-2pt]{0pt}{11pt}\textbf{Training-Free}} \\
ScaleCrafter~\citep{he2024scalecrafter} & 2.37 & 21.17 & 59.61 & 42.13 & 44.87 & 30.01 & 35.06 & 20.40 & 34.61 \\
DemoFusion~\citep{du2024demofusion} & 6.06 & 21.95 & 53.54 & 31.23 & 35.21 & 34.17 & 34.25 & 22.73 & 31.55 \\
HiDiffusion~\citep{zhang2024hidiffusion} & 3.13 & 21.45 & 51.34 & 35.90 & 45.69 & 35.72 & 34.47 & 20.57 & 34.46 \\
FouriScale~\citep{huang2024fouriscale} & 2.40 & 21.19 & 58.09 & 43.93 & 47.48 & 30.83 & 33.56 & 19.66 & 35.22 \\
DiffuseHigh~\citep{kim2025diffusehigh} & 7.43 & 22.23 & 56.37 & 43.58 & 47.95 & 39.73 & 35.08 & 24.88 & 38.32 \\
FreCaS~\citep{zhang2024frecas} & 8.40 & 22.24 & 61.41 & 46.54 & 50.77 & 40.19 & 35.57 & 29.37 & 40.56 \\
AccDiffusion~\citep{lin2025accdiffusion} & 6.92 & 22.03 & 55.12 & 36.51 & 41.35 & 40.02 & 33.89 & 18.96 & 34.19 \\
FreeScale~\citep{qiu2025freescale} & 8.69 & 22.16 & 56.26 & 45.82 & 50.29 & 40.95 & 34.93 & 28.22 & 40.14 \\
HiFlow~\citep{bu2026hiflow} & 11.18 & 22.82 & 58.69 & 47.96 & 56.07 & 48.37 & 36.62 & 35.24 & 44.86 \\
DyPE~\citep{issachar2026dype} & -5.33 & 19.90 & 46.10 & 27.41 & 26.40 & 21.25 & 32.68 & 13.76 & 24.38 \\
ScaleDiff~\citep{koh2026scalediff} & 11.01 & 22.71 & 58.20 & 44.96 & 52.28 & 47.38 & 35.70 & 33.06 & 42.68 \\
SEGA~\citep{rajabi2026sega} & 1.53 & 21.12 & 48.83 & 32.09 & 42.71 & 36.86 & 34.58 & 20.14 & 33.39 \\
SigMa~\citep{zhaosigma} & 2.38 & 21.07 & 58.49 & 36.92 & 38.06 & 32.56 & 34.91 & 21.37 & 32.70 \\
HRDiT~\citep{xue2026hrdit} & 11.36 & 22.85 & 60.48 & 47.37 & 56.11 & 48.15 & 38.14 & 37.08 & 45.36 \\
\addlinespace[2pt]
\rowcolor{black!6}
\multicolumn{10}{@{}l@{}}{\hspace{4pt}\rule[-2pt]{0pt}{11pt}\textbf{Training-Based}} \\
UltraPixel~\citep{ren2024ultrapixel} & 9.23 & 22.68 & 60.70 & 52.18 & 54.92 & 46.03 & 36.84 & 30.29 & 44.23 \\
Sana~\citep{xie2024sana} & 9.81 & 22.48 & 55.20 & 47.08 & 54.20 & 46.52 & 36.39 & 32.63 & 43.42 \\
URAE~\citep{yu2025ultra} & 8.61 & 22.01 & 56.05 & 47.07 & 49.69 & 46.28 & 37.80 & 28.31 & 41.87 \\
Diffusion-4K~\citep{zhang2025diffusion} & 4.30 & 21.20 & 53.61 & 46.74 & 48.91 & 38.93 & 37.14 & 21.86 & 38.93 \\
UltraFlux~\citep{ye2026ultraflux} & 10.52 & 22.51 & \second{62.30} & 50.78 & 54.14 & 47.25 & 37.49 & 36.66 & 45.22 \\
FLUX-3-Image-Klein~\citep{flux-3-image-klein-2026} & 10.93 & 23.02 & 58.37 & \second{54.57} & 59.25 & 60.08 & \second{43.89} & 47.73 & 53.12 \\
FLUX-3-Image-Klein-Base~\citep{flux-3-image-klein-2026} & 10.62 & 22.93 & 56.75 & 53.84 & 58.46 & 59.76 & \best{45.88} & \second{49.91} & \second{53.66} \\
\addlinespace[2pt]
\rowcolor{black!6}
\multicolumn{10}{@{}l@{}}{\hspace{4pt}\rule[-2pt]{0pt}{11pt}\textbf{Same-Backbone Comparison}} \\
Qwen-Image 2.0 SFT & \second{12.27} & \second{23.12} & 60.08 & 54.21 & \second{61.65} & \best{62.19} & 40.58 & 48.33 & 53.40 \\
\textbf{Qwen-Image-HD} & \best{12.36} & \best{23.17} & \best{62.81} & \best{58.05} & \best{64.31} & \second{60.59} & 41.82 & \best{51.80} & \best{55.32} \\
\bottomrule[0.8pt]
\end{tabular}
\endgroup
\end{table}

\begin{figure}[!htbp]
\centering
\includegraphics[width=\linewidth]{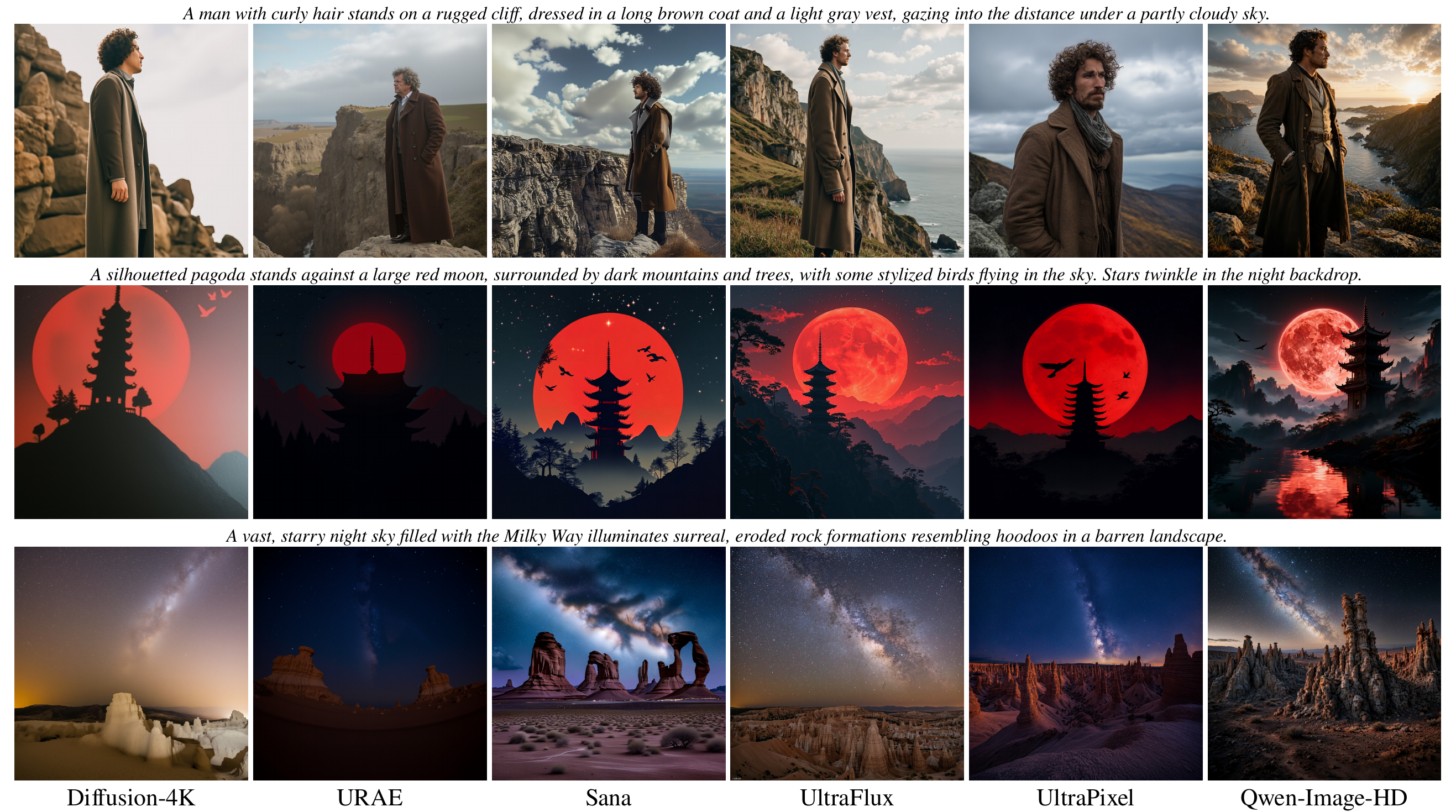}
\caption{\textbf{Generation comparisons on Aesthetic-4K.}}
\label{fig:aes4k}
\end{figure}

\textbf{Qualitative results.}
\Cref{fig:aes4k} compares our method with five training-based external methods. \ours{} preserves scene depth and structural continuity while producing rich local detail, with texture directions and distributions that agree with object shapes.

\begin{figure}[!tbp]
\centering
\includegraphics[width=\linewidth]{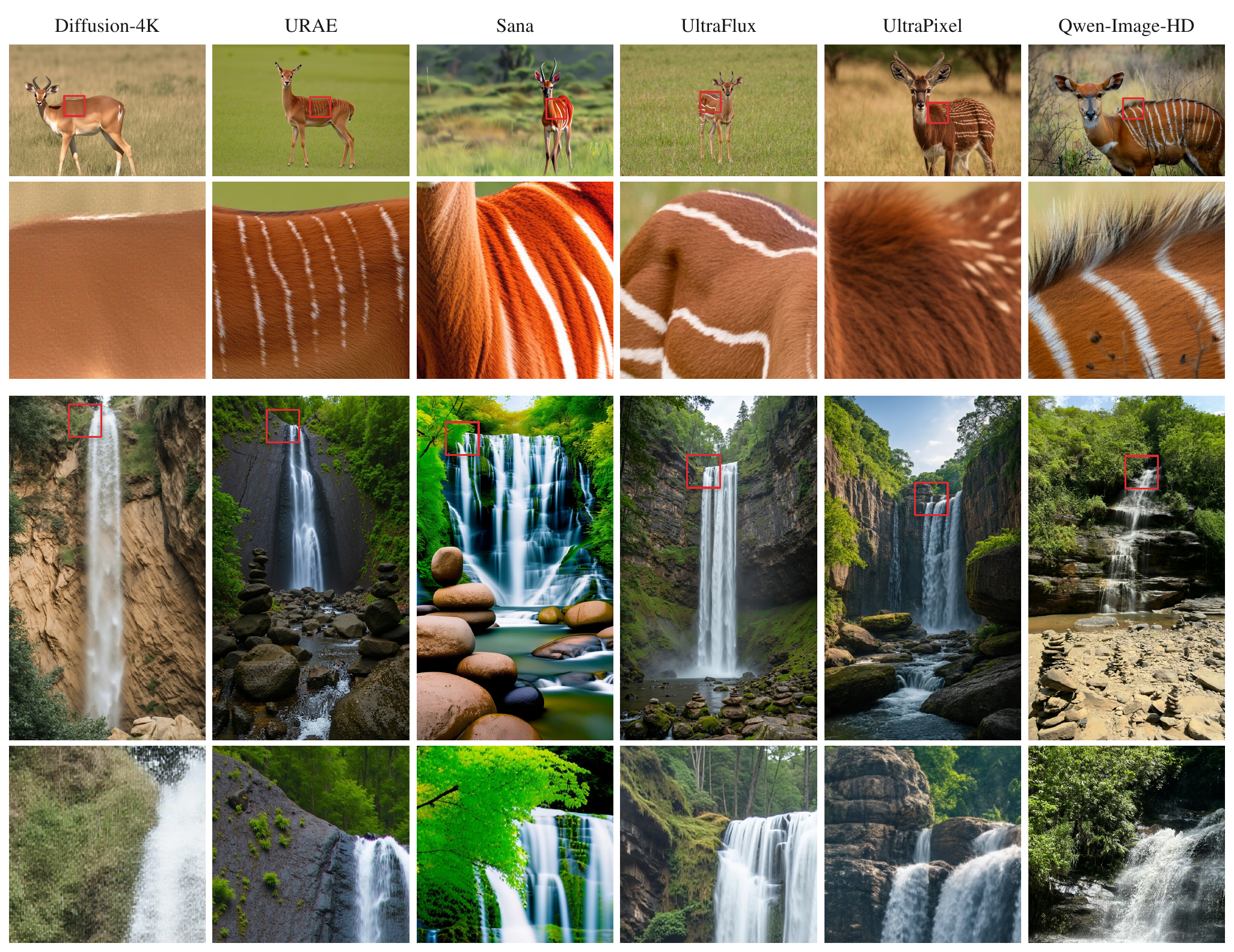}
\caption{\textbf{Texture detail comparisons with five training-based external methods.} Each pair of rows corresponds to one prompt: the upper row shows full images with red crop boxes, and the lower row shows the corresponding enlarged, original-resolution details. Each column presents one method.}
\label{fig:external-texture}
\end{figure}

\Cref{fig:external-texture} further examines local material appearance through original-resolution crops from two additional prompts, compared with the same five external methods. In these examples, \ours{} preserves fine, irregular structures and local texture variation, while several external outputs appear smoother or exhibit overly regular patterns.

\subsection{Training Efficiency and Generation Quality}
\label{sec:training-results}

Our same-backbone comparison examines training cost, Q-Judger Overall, and unnatural repetitive stripes using a spectral diagnostic, complemented by independent human preferences for visual quality and detail clarity.

\textbf{Nature-384 benchmark.}
\label{sec:nature384}
Unnatural repetitive stripes are particularly visible in grass and mountain vegetation, imposing similar fine patterns on different materials. We construct \textbf{Nature-384} with 384 natural-scene text descriptions, generate outputs from the same descriptions for all methods, and examine these patterns at the original resolution.

\textbf{Diagnosing unnatural repetitive stripes with NSE.}
\label{sec:nse}
Periodic stripes concentrate energy in narrow bands of a local spectrum. Based on this property, we define \emph{Narrowband Spectral Excess} (NSE), which measures the excess response of the power spectrum above a local background. It depends only on the generated image and requires neither a reference image nor a trained evaluation model.

We normalize 8-bit RGB code values to $[0,1]$ and compute grayscale code values as $Y=0.2126I_R+0.7152I_G+0.0722I_B$. Windows of size $L=256$ and stride 128 cover the image, with boundary windows aligned to its edges. For each window $Y_j$, we remove low-frequency content and apply a two-dimensional Hann window $h$ to obtain the power spectrum
\begin{equation}
D_j=Y_j-G_\sigma*Y_j,\qquad
P_j(\mathbf k)=\left|\mathcal F\{h\odot D_j\}(\mathbf k)\right|^2,
\qquad \sigma=4.
\label{eq:nse-spectrum}
\end{equation}
Here, $G_\sigma$ is a Gaussian kernel with standard deviation $\sigma$ pixels, $h$ is the outer product of one-dimensional Hann windows, $\mathcal F$ is the unnormalized two-dimensional discrete Fourier transform, and $\mathbf k$ indexes frequency bins relative to zero frequency. High-pass filtering suppresses slow luminance variations, and the Hann window reduces spectral leakage. We evaluate the band $\Omega=\{\mathbf k:L/32\leq\|\mathbf k\|\leq L/2\}$, corresponding to frequencies from $1/32$ to $1/2$ cycles/pixel.

To distinguish narrowband peaks from broadband responses caused by natural edges, we estimate the background using both a two-dimensional neighborhood and a radial neighborhood along the same direction. Let $B_j^{\mathrm{2D}}$ be the $9\times9$ neighborhood median, $\rho=\|\mathbf k\|$, and $\mathbf e=\mathbf k/\rho$. We define
\begin{equation}
\begin{aligned}
B_j^{\mathrm{rad}}(\mathbf k)
&=\operatorname{median}_{\delta\in\mathcal D}
\widetilde P_j\!\left(\max\{\rho+\delta,1\}\,\mathbf e\right),\\
B_j(\mathbf k)
&=\max\left\{B_j^{\mathrm{2D}}(\mathbf k),B_j^{\mathrm{rad}}(\mathbf k)\right\},
\end{aligned}
\quad \mathcal D=\{\pm3,\pm4,\pm5,\pm6,\pm7,\pm8\}.
\label{eq:nse-background}
\end{equation}
Offsets $\delta$ are measured in frequency bins. $\widetilde P_j$ uses bilinear interpolation with periodic boundaries; Gaussian and two-dimensional median filtering use reflected boundaries. The radial estimate accounts for energy variation along the same direction, suppressing directional broadband responses. Taking the larger background retains narrowband peaks that exceed both estimates.

Let $Z_h=L^2\sum_{\mathbf u}h(\mathbf u)^2$, where $\mathbf u$ ranges over spatial window positions. Band energy, excess energy, and the window score are
\begin{equation}
\begin{aligned}
E_j&=\frac{1}{Z_h}\sum_{\mathbf k\in\Omega}P_j(\mathbf k),\\
E_j^{\mathrm{ex}}&=\frac{1}{Z_h}\sum_{\mathbf k\in\Omega}
\left[P_j(\mathbf k)-\lambda B_j(\mathbf k)\right]_+,\\
s_j&=\frac{E_j^{\mathrm{ex}}}{E_j+\tau^2}.
\end{aligned}
\label{eq:nse-window}
\end{equation}
We use $[x]_+=\max(x,0)$, $\lambda=6$, and $\tau=4/255$, with $0\leq s_j\leq1$. The factor $Z_h$ compensates for Fourier and Hann-window energy scaling. The regularizer $\tau^2$ continuously downweights weak spectral peaks in low-contrast regions, preventing smooth areas such as sky from receiving high scores solely because their energy is relatively concentrated.

We average scores from overlapping windows at each pixel and compute the full-image score:
\begin{equation}
A_I(\mathbf p)=\frac{1}{|\mathcal W(\mathbf p)|}\sum_{j\in\mathcal W(\mathbf p)}s_j,
\qquad
\operatorname{NSE}(I)=\frac{1}{HW}\sum_{\mathbf p}A_I(\mathbf p).
\label{eq:nse-image}
\end{equation}
Here, $\mathcal W(\mathbf p)$ contains windows covering pixel $\mathbf p$, and $H,W$ are the original image dimensions. Dataset scores average image-level NSE with equal weight. Lower NSE indicates weaker narrowband excess associated with unnatural repetitive stripes.

To validate NSE against human judgments of unnatural repetitive stripes, we randomly sample Nature-384 cases and extract $1024\times1024$ crops at matching, randomly selected positions from \ours{} and full-image SFT outputs. Four human annotators evaluate 400 samples. The crops are presented individually in randomized order, with method identities and NSE scores hidden, and rated on a five-point scale for the severity of unnatural repetitive stripes, from absent to severe. Among matched pairs receiving different human scores, crop-level NSE agrees with human ordering in 90.79\% of comparisons (95\% CI: 83.58\%--97.01\%), supporting its consistency with the perceived severity of unnatural repetitive stripes. The confidence interval is estimated using 2,000 bootstrap resamples of original cases, keeping the two method crops paired.

\textbf{Training budget and quality.}
Training compute is estimated from the generator's forward and backward FLOPs, averaged over the training distribution of image shapes, text lengths, and historical conditions, and reported cumulatively in EFLOPs ($10^{18}$ floating-point operations). With the same global batch size and GPU allocation, \ours{} and SFT average approximately 2.7\,s and 41.5\,s per training step, respectively, yielding a $15.4\times$ speedup. The shorter image sequence described in \cref{sec:history} lowers per-step cost, allowing more updates within a fixed training-time budget.

\begin{figure}[!tbp]
\centering
\includegraphics[width=0.85\linewidth]{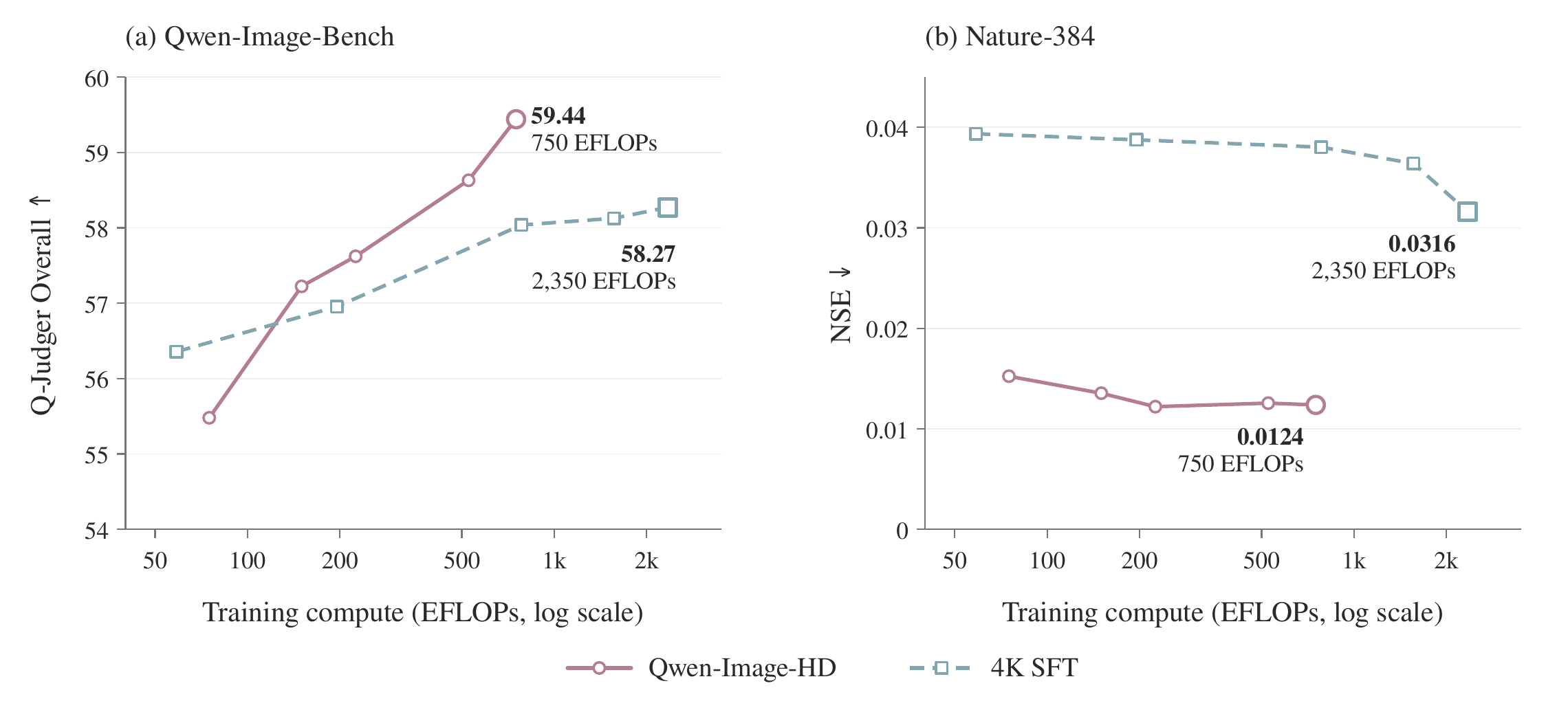}
\caption{\textbf{Training compute and generation quality of \ours{} and full-image 4K SFT.} Both methods use the same backbone. (a) Q-Judger Overall on Qwen-Image-Bench and (b) NSE on Nature-384 versus estimated cumulative Transformer training FLOPs. Lower NSE indicates weaker narrowband spectral responses associated with unnatural repetitive stripes.}
\label{fig:training}
\end{figure}

Within the evaluated training budgets and configurations, \ours{} obtains 59.44 Q-Judger Overall with 750 EFLOPs, compared with 58.27 for SFT using 2,350 EFLOPs: a 68.0\% reduction in estimated training FLOPs and a 1.17-point Overall improvement. It initially trails SFT while adapting to blockwise generation, but improves with fluctuations as training proceeds. Already at 525 EFLOPs, it scores 58.63, exceeding the highest SFT score shown in \cref{fig:training}(a).

\Cref{fig:training}(b) shows that \ours{} maintains lower NSE than SFT throughout the displayed training range. At each method's highest-Overall checkpoint in this range, mean NSE is 0.0124 for \ours{} versus 0.0316 for SFT, a 60.85\% reduction. This gap is consistent with the more pronounced unnatural repetitive stripes visible in the SFT crops.

\begin{figure}[!tbp]
\centering
\includegraphics[width=\linewidth]{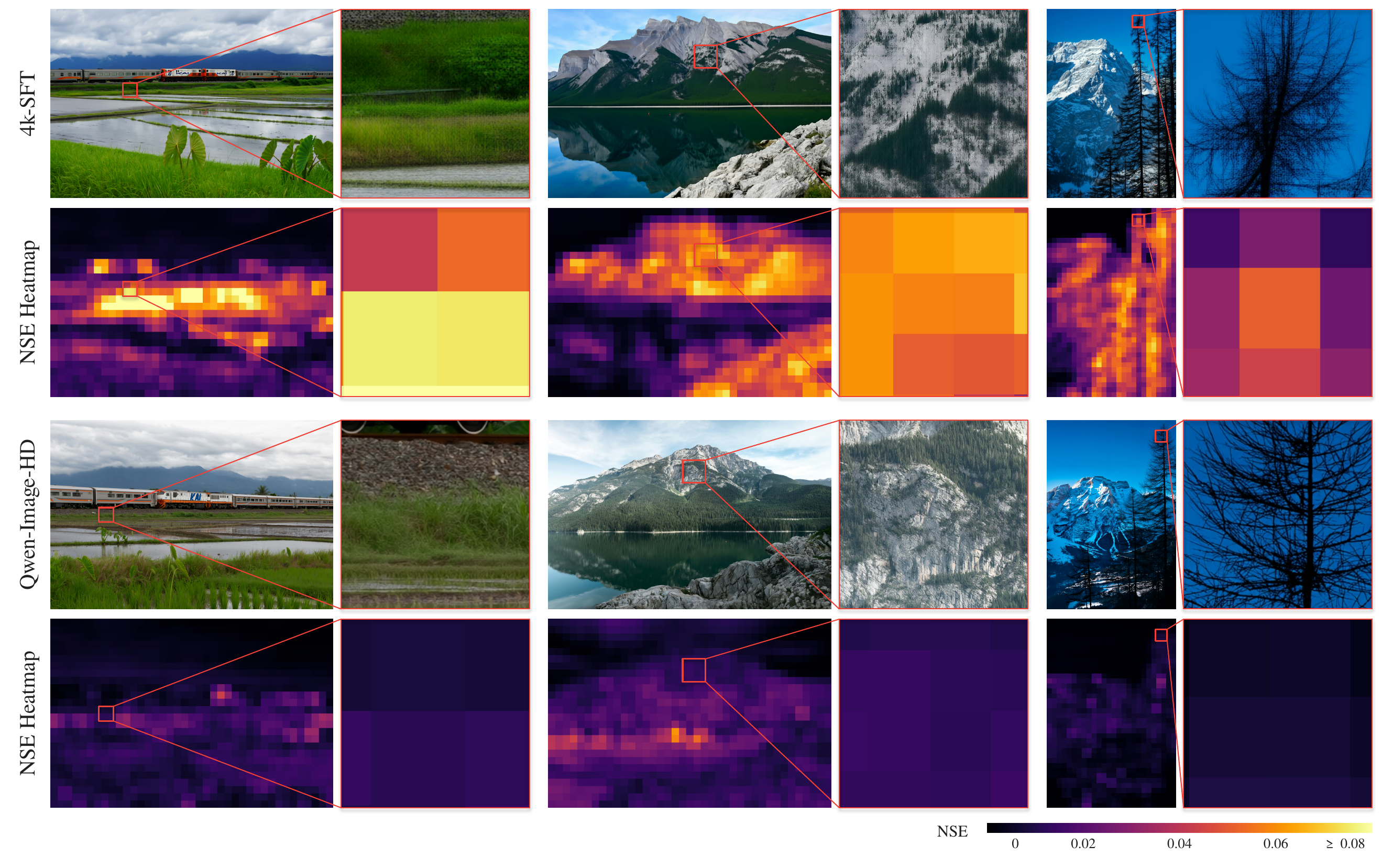}
\caption{\textbf{Unnatural repetitive stripes and NSE heatmaps.} Red boxes mark corresponding crops; heatmaps share one color scale.}
\label{fig:texture}
\end{figure}

\textbf{Original-resolution texture and materials.}
\Cref{fig:texture} pairs original-resolution crops of grass, mountain vegetation, and branches with NSE heatmaps. Unnatural repetitive stripes recur across SFT regions alongside stronger narrowband responses. In comparison, crops from \ours{} show visible contours, scale variation, and irregular details with less pronounced repetitive stripes.

\par\medskip
\noindent
\begin{minipage}[b]{0.56\linewidth}
\textbf{Human preference evaluation.} Since NSE does not measure detail clarity, four human annotators evaluate 400 examples, judging visual quality and detail clarity separately. \Cref{fig:human-preference} shows that \ours{} is preferred over SFT on both dimensions: 53.3\% versus 26.4\% for visual quality, and 68.3\% versus 7.2\% for detail clarity.
\end{minipage}\hfill
\begin{minipage}[b]{0.42\linewidth}
\centering
\includegraphics[width=\linewidth]{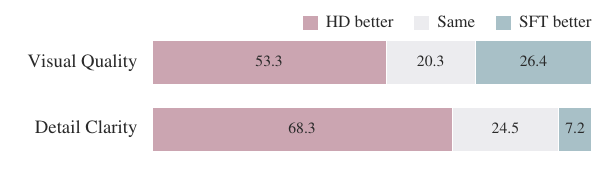}
\captionsetup{font=footnotesize,skip=4pt,hypcap=false}
\captionof{figure}{\textbf{Human preferences for \ours{} versus SFT.}}
\label{fig:human-preference}
\end{minipage}
\par\medskip

\subsection{Ablation Studies}
\label{sec:ablations}

Seven variants examine the roles of the historical thumbnail, neighboring tiles, spatial alignment, and neighbor extent. The full model uses upper and left neighbors with a depth of 256 pixels. Each variant is trained independently from the same backbone with matched training data, learning rate, random seed, and global target-tile batch size, and evaluated on Qwen-Image-Bench.

\begin{table}[!tbp]
\centering
\caption{\textbf{Ablations on Qwen-Image-Bench.} Bold and underlining denote the best and second-best scores.}
\label{tab:ablation}
\begingroup
\fontsize{8.5}{10.2}\selectfont
\setlength{\tabcolsep}{0pt}
\renewcommand{\arraystretch}{1.15}
\setlength{\aboverulesep}{3pt}
\setlength{\belowrulesep}{3pt}
\begin{tabular}{@{}>{\raggedright\arraybackslash}m{.318\linewidth}*{6}{>{\centering\arraybackslash}m{.113666\linewidth}}@{}}
\toprule[0.8pt]
{\fontsize{8}{9.5}\selectfont \textbf{Variant}} & \multicolumn{1}{c}{{\fontsize{8}{9.5}\selectfont \textbf{Quality}\,$\uparrow$}} & \multicolumn{1}{c}{{\fontsize{8}{9.5}\selectfont \textbf{Aesthetics}\,$\uparrow$}} & \multicolumn{1}{c}{{\fontsize{8}{9.5}\selectfont \textbf{Alignment}\,$\uparrow$}} & \multicolumn{1}{c}{{\fontsize{8}{9.5}\selectfont \shortstack[c]{\textbf{Real-World}\\\textbf{Fidelity}}\,$\uparrow$}} & \multicolumn{1}{c}{{\fontsize{8}{9.5}\selectfont \shortstack[c]{\textbf{Creative}\\\textbf{Generation}}\,$\uparrow$}} & \multicolumn{1}{c}{{\fontsize{8}{9.5}\selectfont \textbf{Overall}\,$\uparrow$}} \\
\midrule[0.4pt]
\mbox{\textbf{Full Model}} & \best{56.99} & \best{65.79} & \best{60.42} & \best{47.59} & \best{60.59} & \best{59.44} \\
\addlinespace[3pt]
\mbox{Recurrent Memory} & 55.68 & 62.02 & 56.31 & 45.26 & 54.40 & 56.10 \\
\addlinespace[3pt]
\mbox{Learned Compression} & 55.60 & 63.02 & 57.11 & 45.97 & 56.32 & 56.88 \\
\addlinespace[3pt]
\mbox{w/o Historical Thumbnail} & 52.46 & 52.27 & 50.63 & 45.05 & 48.70 & 50.49 \\
\addlinespace[3pt]
\mbox{w/o Neighbor Tiles} & 55.52 & 63.18 & 58.37 & \second{47.40} & 56.76 & 57.36 \\
\addlinespace[3pt]
\mbox{w/o Thumbnail--Canvas Alignment} & 55.08 & 63.43 & 58.51 & 46.20 & 57.46 & 57.38 \\
\addlinespace[3pt]
\mbox{w/o Neighbor--Target Alignment} & 55.78 & 63.59 & 58.00 & 46.34 & 57.62 & 57.41 \\
\addlinespace[3pt]
\mbox{Full-size Neighbor Tiles} & \second{56.82} & \second{65.54} & \second{60.14} & 46.73 & \second{59.76} & \second{59.07} \\
\bottomrule[0.8pt]
\end{tabular}
\endgroup
\end{table}

\begin{figure}[!tbp]
\centering
\includegraphics[width=\linewidth]{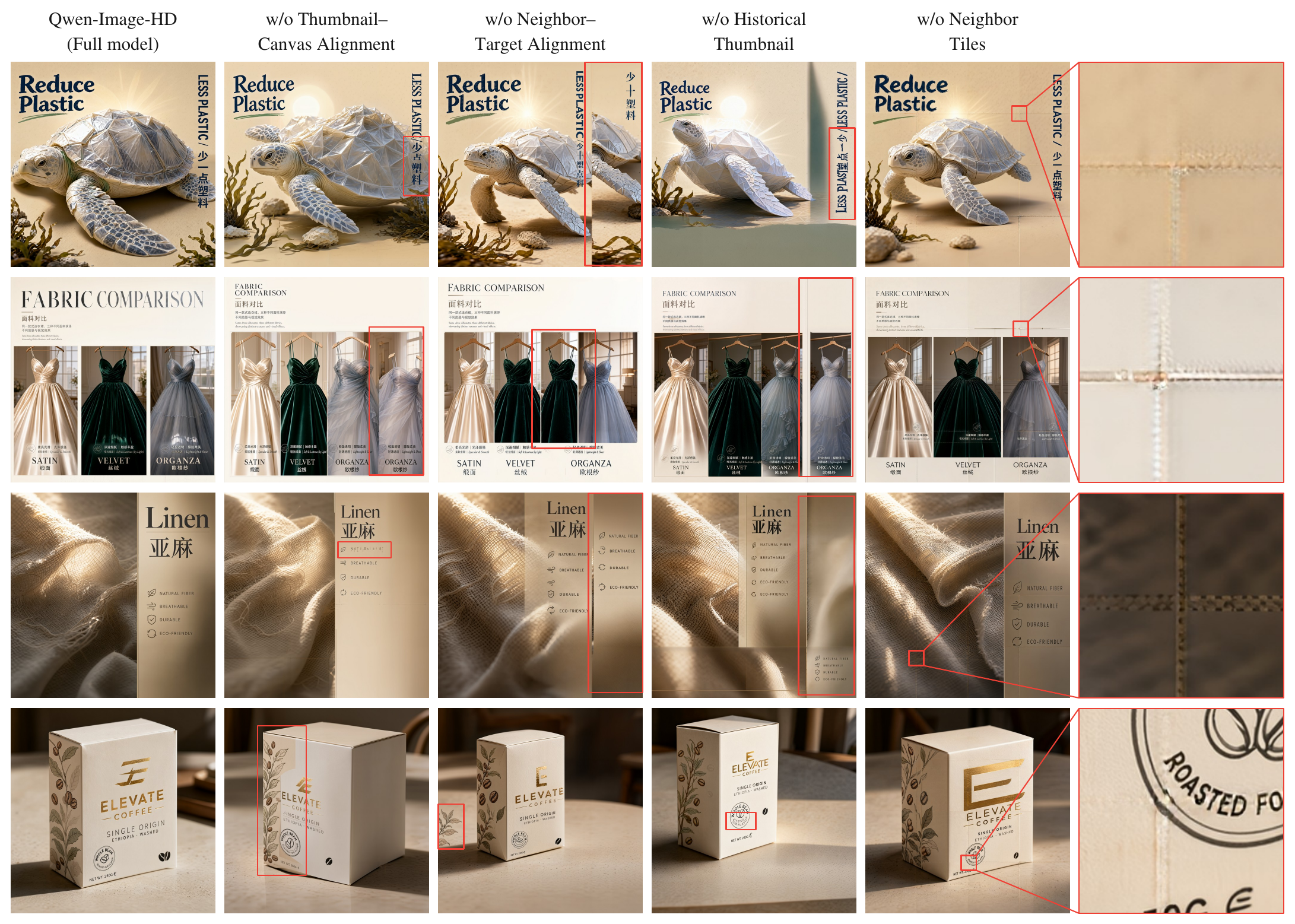}
\caption{\textbf{Visual ablations of historical conditions and spatial alignment.} Character arrangement and overall typography in these text-oriented cases require coordination across tiles, exposing changes in global layout and consistency. The first five columns show the full model, removal of thumbnail--canvas alignment, removal of neighbor--target alignment, removal of the historical thumbnail, and removal of neighbors. The final column enlarges the red-boxed region of the no-neighbor result.}
\label{fig:ablation}
\end{figure}

\textbf{Historical thumbnail.}
Removing the thumbnail produces the largest Overall decrease in \cref{tab:ablation}, 8.95 points, with Aesthetics and Alignment falling by 13.53 and 9.79 points. Despite retaining high-resolution neighbors, this variant exhibits abrupt region transitions and repeated local content in \cref{fig:ablation}, indicating that neighboring detail alone is insufficient for coherent organization across regions.

\textbf{Alternative history representations.}
Recurrent Memory and Learned Compression replace only the historical-thumbnail token block while retaining the backbone, 256-pixel neighbors, shared canvas coordinates, training objective, and the same history-token budget. Completed latent tiles are independently average-pooled into corresponding cells of a compact canvas; target and future tiles are excluded. Each history encoder receives these values, an availability mask, and normalized spatial coordinates, and outputs the baseline thumbnail geometry and token count: $64\times64$ tokens for a square 4K image. Both encoders use 96 hidden channels and two residual spatial blocks with depthwise $3\times3$ and pointwise convolutions, followed by projection to the original latent channels. Unobserved cells use a learned empty embedding, and the encoders are trained jointly with the generator.

Inspired by NUWA-Infinity's~\citep{liang2022nuwa} neighborhood-based history reuse, Recurrent Memory merges upper/left predecessor states using learned gates and updates the merged state with the current pooled tile. The target receives the merged states of its upper/left neighbors. Training reconstructs the causal recurrence from preceding ground-truth tiles; Serial inference updates it after each tile and discards states not needed by later neighbors. Learned Compression draws on MADFormer's~\citep{chen2025madformer} separation of autoregressive conditioning and diffusion refinement, using a lightweight encoder to construct history before iterative target denoising. The encoder predicts a residual over pooled historical latents in observed cells. The pooled canvas is updated after each completed tile, and the condition is recomputed for the next target.

Their Overall scores are 56.10 and 56.88, respectively, compared with 59.44 for the full model (\cref{tab:ablation}). These controlled alternatives test adapted history representations within our framework, supporting the thumbnail among the compared designs.

\textbf{High-resolution neighbors.}
Without neighbors, the thumbnail still establishes a coarse layout, but seams appear at tile boundaries, forming crosses where horizontal and vertical boundaries meet, together with brightness or texture discontinuities (\cref{fig:ablation}).

\textbf{Spatial alignment.}
Independently centering the thumbnail, or independently centering the neighbors and target, reduces Overall by 2.06 and 2.03 points, respectively. Both retain the same conditions and tile count while changing position encodings, and both decrease all five dimension scores. The former disrupts correspondence between coarse history and the target region; the latter weakens relative positions between adjacent tiles. Repeated subjects and displaced layouts in \cref{fig:ablation} reflect these losses.

\textbf{Neighbor size.}
The Full-size Neighbor Tiles variant retains the same upper and left neighbors but expands each to a complete $1024\times1024$ tile. This yields an Overall score of 59.07, 0.37 points below the 256-pixel-neighbor full model; all five dimension scores are also lower. Larger neighbors therefore bring no quality gain in this comparison, while requiring longer condition sequences and more computation.

\subsection{Wavefront Inference Time and Generation Quality}
\label{sec:inference-results}

We compare full-image SFT, Serial, Basic Wavefront, and optimized Wavefront using four GPUs. The primary Serial and Wavefront comparison shares \ours{} weights from Serial training, generation prompts, sampling settings, and random seeds. We also evaluate a Wavefront-trained variant to examine the effect of matching training and inference history. \Cref{tab:wavefront} reports mean latency and Q-Judger scores on Qwen-Image-Bench, and NSE on Nature-384; Training Strategy identifies the history visibility used during training.

\begin{table}[!tbp]
\centering
\caption{\textbf{4K inference efficiency and generation quality.} Training Strategy specifies the history used during training. For SFT and Serial, the symbol ``/'' separates one- and four-GPU results. Basic uses one GPU per tile; Opt. adds CFG/SP and GPU reassignment. Q-Judger is evaluated on Qwen-Image-Bench and NSE on Nature-384. Bold denotes the best entries.}
\label{tab:wavefront}
\begingroup
\fontsize{8}{10}\selectfont
\setlength{\tabcolsep}{0pt}
\renewcommand{\arraystretch}{1.15}
\setlength{\aboverulesep}{2pt}
\setlength{\belowrulesep}{2pt}
\begin{tabular}{@{}>{\raggedright\arraybackslash}m{0.1730\linewidth}>{\centering\arraybackslash}m{0.0820\linewidth}>{\centering\arraybackslash}m{0.0420\linewidth}>{\centering\arraybackslash}m{0.0830\linewidth}>{\centering\arraybackslash}m{0.0910\linewidth}>{\centering\arraybackslash}m{0.0765\linewidth}>{\centering\arraybackslash}m{0.0765\linewidth}>{\centering\arraybackslash}m{0.0765\linewidth}>{\centering\arraybackslash}m{0.0765\linewidth}>{\centering\arraybackslash}m{0.0765\linewidth}>{\centering\arraybackslash}m{0.0765\linewidth}>{\centering\arraybackslash}m{0.0700\linewidth}@{}}
\toprule[0.8pt]
\multirow{3}{*}[-3pt]{\textbf{Method}} & \multirow{3}{*}[-3pt]{\shortstack{\textbf{Training}\\\textbf{Strategy}}} & \multicolumn{3}{c}{\textbf{Inference Efficiency}} & \multicolumn{6}{c}{\textbf{Q-Judger}} & \multicolumn{1}{c}{\multirow{3}{*}[-3pt]{\textbf{NSE} $\downarrow$}} \\
\cmidrule(lr){3-5}\cmidrule(lr){6-11}
 & & \multicolumn{1}{c}{{\fontsize{6.5}{8.5}\selectfont\textit{GPUs/}}} & \multicolumn{1}{c}{{\fontsize{6.5}{8.5}\selectfont\textit{Time}}} & \multicolumn{1}{c}{{\fontsize{6.5}{8.5}\selectfont\textit{Images/}}} & \multicolumn{1}{c}{\multirow{2}{*}{{\fontsize{6.5}{8.5}\selectfont\textit{Quality}\,$\uparrow$}}} & \multicolumn{1}{c}{\multirow{2}{*}{{\fontsize{6.5}{8.5}\selectfont\textit{Aesthetics}\,$\uparrow$}}} & \multicolumn{1}{c}{\multirow{2}{*}{{\fontsize{6.5}{8.5}\selectfont\textit{Alignment}\,$\uparrow$}}} & \multicolumn{1}{c}{{\fontsize{6.5}{8.5}\selectfont\textit{Real-World}}} & \multicolumn{1}{c}{{\fontsize{6.5}{8.5}\selectfont\textit{Creative}}} & \multicolumn{1}{c}{\multirow{2}{*}{{\fontsize{6.5}{8.5}\selectfont\textit{Overall}\,$\uparrow$}}} &  \\
 & & \multicolumn{1}{c}{{\fontsize{6.5}{8.5}\selectfont\textit{Image}}} & \multicolumn{1}{c}{{\fontsize{6.5}{8.5}\selectfont\textit{(s) $\downarrow$}}} & \multicolumn{1}{c}{{\fontsize{6.5}{8.5}\selectfont\textit{GPU$\cdot$h $\uparrow$}}} &  &  &  & \multicolumn{1}{c}{{\fontsize{6.5}{8.5}\selectfont\textit{Fidelity}\,$\uparrow$}} & \multicolumn{1}{c}{{\fontsize{6.5}{8.5}\selectfont\textit{Generation}\,$\uparrow$}} &  &  \\
\midrule[0.4pt]
\shortstack[l]{Qwen-Image 2.0\\SFT} & --- & \mbox{1\,/\,4} & \mbox{660\,/\,180} & \mbox{5.45\,/\,4.99} & 55.25 & 64.54 & 59.35 & 46.49 & 60.20 & 58.27 & 0.0316 \\
\addlinespace[4pt]
\shortstack[l]{Qwen-Image-HD\\Serial} & Serial & \mbox{1\,/\,4} & \mbox{520\,/\,166} & \mbox{\best{6.92}\,/\,5.44} & \best{56.99} & \best{65.79} & \best{60.42} & \best{47.59} & \best{60.59} & \best{59.44} & 0.0124 \\
\addlinespace[4pt]
\shortstack[l]{Qwen-Image-HD\\Wavefront (Basic)} & Serial & 4 & 228 & 3.95 & 56.13 & 65.73 & 58.80 & 47.22 & 58.90 & 58.52 & \best{0.0123} \\
\addlinespace[4pt]
\shortstack[l]{Qwen-Image-HD\\Wavefront (Opt.)} & Serial & 4 & \best{139} & 6.45 & 56.13 & 65.73 & 58.80 & 47.22 & 58.90 & 58.52 & \best{0.0123} \\
\addlinespace[4pt]
\shortstack[l]{Qwen-Image-HD\\Wavefront (Opt.)} & Wavefront & 4 & \best{139} & 6.45 & 56.54 & 64.62 & 58.24 & 46.88 & 59.38 & 58.26 & 0.0169 \\
\bottomrule[0.8pt]
\end{tabular}
\endgroup
\end{table}

\textbf{Inference efficiency.}
Using the generation settings above, timing includes condition preparation, sampling, communication, history updates, and decoding, and excludes model loading, warmup, and PNG saving. For latency $T$ seconds with $G$ collaborating GPUs, we report $3600/(GT)$ images per GPU-hour to reflect GPU resource use.

Optimized Wavefront (\cref{fig:wavefront}(c)) reduces mean latency from Basic's 228 seconds to 139 seconds, an approximately 39\% reduction. With four GPUs per image, it also reduces latency by approximately 16\% versus Serial's 166 seconds and 23\% versus SFT's 180 seconds. All three four-GPU configurations---full-image SFT, Serial, and optimized Wavefront---use CFG and sequence parallelism.

Under the same four-GPU budget, this latency reduction increases GPU-normalized throughput to 6.45 images per GPU-hour, compared with 5.44 for Serial (\cref{tab:wavefront}).

\textbf{Generation quality.}
Applying Wavefront to the Serial-trained model requires no additional training, but Q-Judger Overall decreases from 59.44 to 58.52, a relative reduction of 1.54\%, while remaining 0.25 points above SFT. Aesthetics changes little, while the largest relative decreases occur in Alignment and Creative Generation, at 2.68\% and 2.79\%, respectively.

\begin{figure}[!htbp]
\centering
\includegraphics[width=\linewidth]{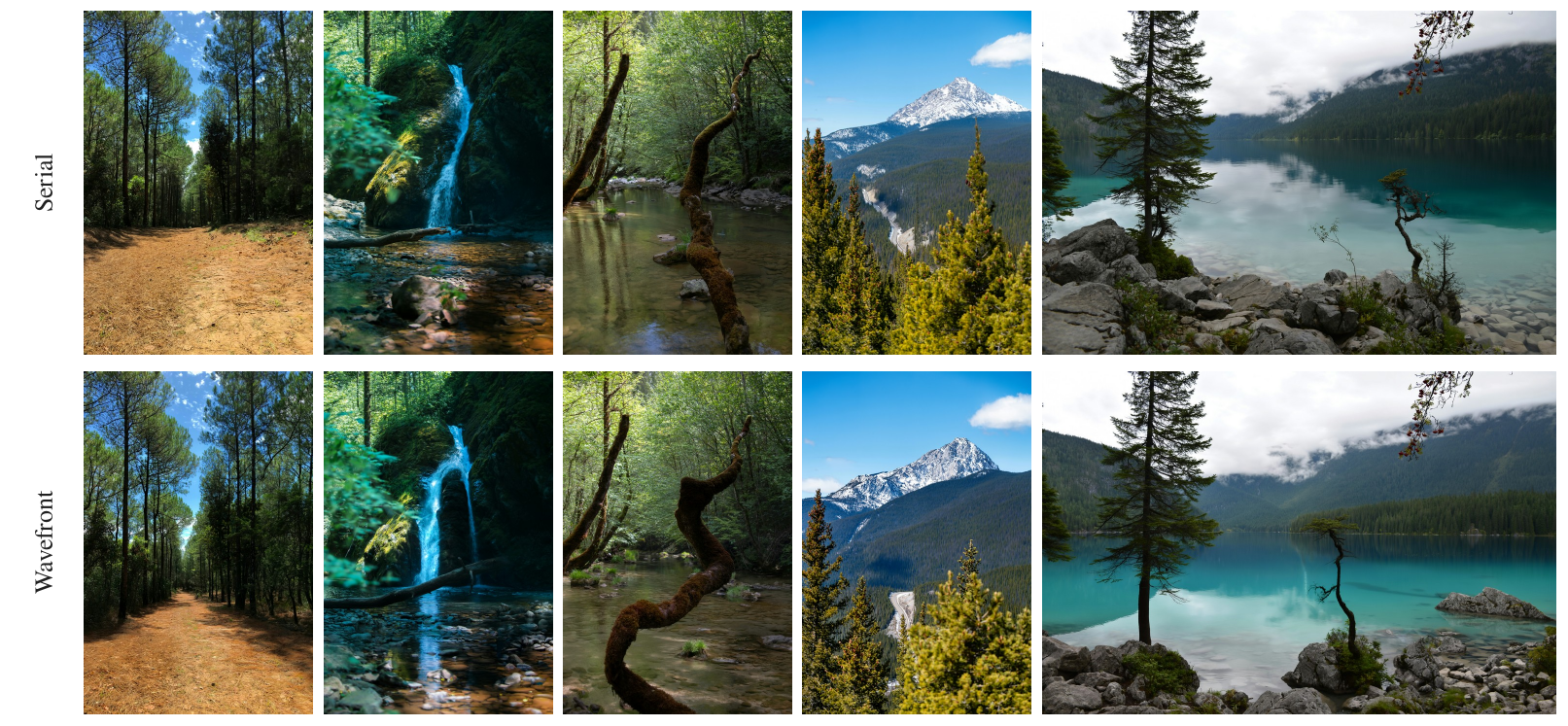}
\caption{\textbf{Serial and Wavefront generation on Nature-384.} The first row shows Serial autoregressive generation, and the second shows Wavefront generation using the same model weights and prompts.}
\label{fig:wavefront-comparison}
\end{figure}

On Nature-384, Wavefront with Serial training achieves an NSE of 0.0123, close to 0.0124 for Serial inference and below SFT's 0.0316. Thus, the two schedules have similar narrowband excess despite their difference in Q-Judger Overall. The visual comparison in \cref{fig:wavefront-comparison} also shows coherent overall scene structure when Wavefront is applied to the Serial-trained model.

\textbf{Training strategy.}
Wavefront training matches history visibility at inference: each tile conditions only on completed preceding rounds. Its Q-Judger Overall of 58.26 does not recover the Serial score of 59.44, suggesting that training--inference mismatch alone does not explain the decrease. Instead, the remaining gap may reflect limited cross-region coordination, as concurrent tiles cannot condition on one another. Its NSE of 0.0169 is also higher than Wavefront with Serial training, although still below full-image SFT.

\FloatBarrier
\section{Conclusion}
\label{sec:conclusion}

We introduced \ours{}, which adapts a pretrained generator through local conditional training and constructs complete images by blockwise autoregression. High\babelhyphen{hard}resolution neighbor boundaries and a historical thumbnail provide complementary scales of context within a bounded token budget, while shared canvas coordinates connect these inputs to the target tile. Experiments on three benchmarks demonstrate strong 4K generation performance. Under the evaluated training budgets and configurations, the same-backbone comparison shows a higher Overall score with lower training compute than full-image SFT. Close-up comparisons show reduced unnatural repetitive stripes, and ablations and controlled history comparisons support the historical thumbnail and spatial alignment within our framework. Wavefront scheduling further accelerates inference through parallel tile generation, offering a latency--quality trade-off versus Serial.

\clearpage
\bibliographystyle{colm2024_conference}
\bibliography{references}
\end{document}